\documentclass[lettersize,journal]{IEEEtran}
\usepackage{algorithmic}
\usepackage{algorithm}
\usepackage{array}
\usepackage[caption=false,font=normalsize,labelfont=sf,textfont=sf]{subfig}
\usepackage{textcomp}
\usepackage{stfloats}
\usepackage{url}
\usepackage{verbatim}
\usepackage{graphicx}
\usepackage{cite}
\usepackage{amsmath,amssymb,amsfonts}
\graphicspath{{figure/}}
\usepackage{xspace}
\usepackage{booktabs}
\usepackage{multirow}
\usepackage{xcolor}
\usepackage{enumitem}
\usepackage{hyperref}

\definecolor{darkblue}{rgb}{0.15,0.15,0.55}
\definecolor{lightgrey}{rgb}{0.75,0.75,0.75}

\newcommand{\nphard}{$\mathcal{NP}$-hard\xspace}

\providecommand{\codecomment}[1]{\textcolor{lightgrey}{\dotfill}\textcolor{darkblue}{//\textrm{#1}}}
\providecommand{\codecommentline}[1]{\textcolor{darkblue}{\textbackslash*\,\textrm{#1}*\textbackslash\,}}

\begin{document}

\title{A Hierarchical Bin Packing Framework with Dual Manipulators via Heuristic Search and Deep Reinforcement Learning}

\author{
Beomjoon Lee$^{1}$,
Changjoo Nam$^{1}$ \\
$^1$Department of Electronic Engineering, Sogang University\\
\texttt{b.lee@sogang.ac.kr},
\texttt{cjnam@sogang.ac.kr}
}



% % column for its text to clear the IEEEpubid mark.

\maketitle

\begin{abstract}
We address the bin packing problem (BPP), which aims to maximize bin utilization when packing a variety of items. The offline problem, where the complete information about the item set and their sizes is known in advance, is proven to be NP-hard. The semi-online and online variants are even more challenging, as full information about incoming items is unavailable. While existing methods have tackled both 2D and 3D BPPs, the 2D BPP remains underexplored in terms of fully maximizing utilization.
We propose a hierarchical approach for solving the 2D online and semi-online BPP by combining deep reinforcement learning (RL) with heuristic search.
The heuristic search selects which item to pack or unpack, determines the packing order, and chooses the orientation of each item,
while the RL agent decides the precise position within the bin.
Our method is capable of handling diverse scenarios, including repacking, varying levels of item information, differing numbers of accessible items, and coordination of dual manipulators.
Experimental results demonstrate that our approach achieves near-optimal utilization across various practical scenarios, largely due to its repacking capability.
In addition, the algorithm is evaluated in a physics-based simulation environment, where execution time is measured to assess its real-world performance.
\end{abstract}

\def\abstractname{Note to Practitioners}
\begin{abstract}

Bin packing is a key task in logistics and manufacturing, where items must be packed tightly into limited space within a short amount of time. Traditional systems typically use a single robot arm and pack items in the order they arrive, without the ability to rearrange previous placements. This limits space use and makes it hard to improve the result once early placements are made.
This work introduces a packing system that incorporates a repacking strategy, allowing previously placed items to be moved when necessary to improve space usage. The system also supports dual-arm operation, where two robot arms work in parallel to reduce overall task time. Items are assigned based on reachability, and placement decisions are made using a combination of predefined rules and AI.
The system is intended for environments such as automated warehouses and production lines, where items arrive one by one and must be packed accurately in real time. It was tested in simulation using UR5e robots and open-source planning tools, demonstrating effective use of bin space and faster task execution with the dual-arm setup compared to the single-arm setup.
\end{abstract}


\section{Introduction}
\label{sec:introduction}

\IEEEPARstart{T}{he} bin packing problem (BPP) has found applications in delivery, warehouse automation, and manufacturing. BPP is a classic combinatorial optimization problem that aims to pack items into a bin while minimizing wasted space. As BPP is \nphard~\cite{NPcompleteness}, various efficient methods have been developed, including heuristics~\cite{2DSurvey, genetic2D, Skyline, ShelfNextFit, 3Dheuristic}. More recently, reinforcement learning (RL) approaches~\cite{deeppack, solving, generalized3D, Recent3DRearrangement} have demonstrated benefits in reducing computational costs.

\begin{figure}[t]
    \centering

    \begin{minipage}{\linewidth}
        \raggedright
        \includegraphics[width=1.0\linewidth]{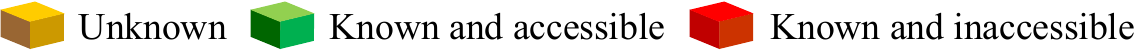}
    \end{minipage}

    \par\medskip

    \begin{minipage}{0.49\columnwidth}
        \centering
        \includegraphics[width=1.0\linewidth]{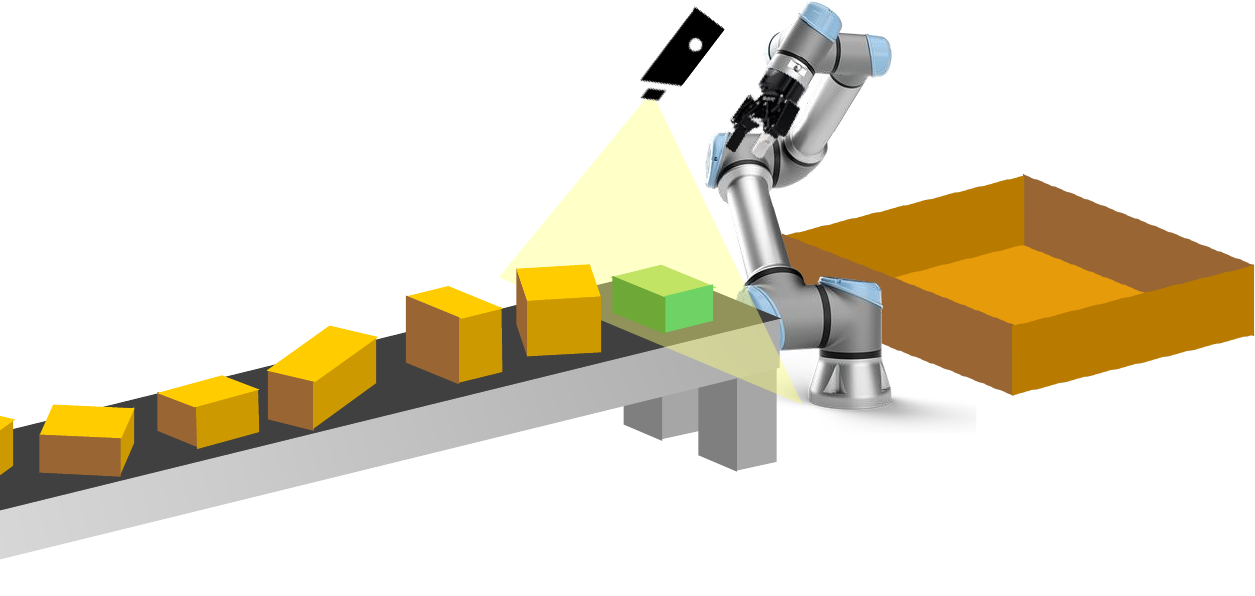}
        
        {\fontsize{8pt}{10pt}\selectfont (a)}
    \end{minipage}%
    \hfill
    \begin{minipage}{0.49\columnwidth}
        \centering
        \includegraphics[width=1.0\linewidth]{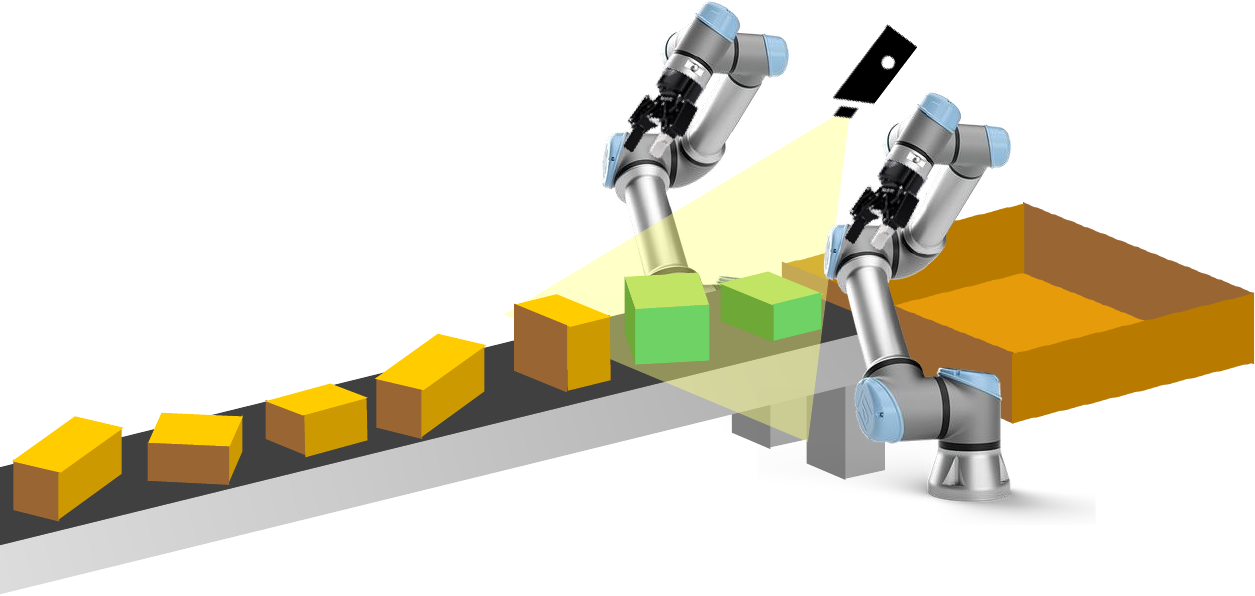}
        
        {\fontsize{8pt}{10pt}\selectfont (d)}
    \end{minipage}

    \par\medskip

    \begin{minipage}{0.49\columnwidth}
        \centering
        \includegraphics[width=1.0\linewidth]{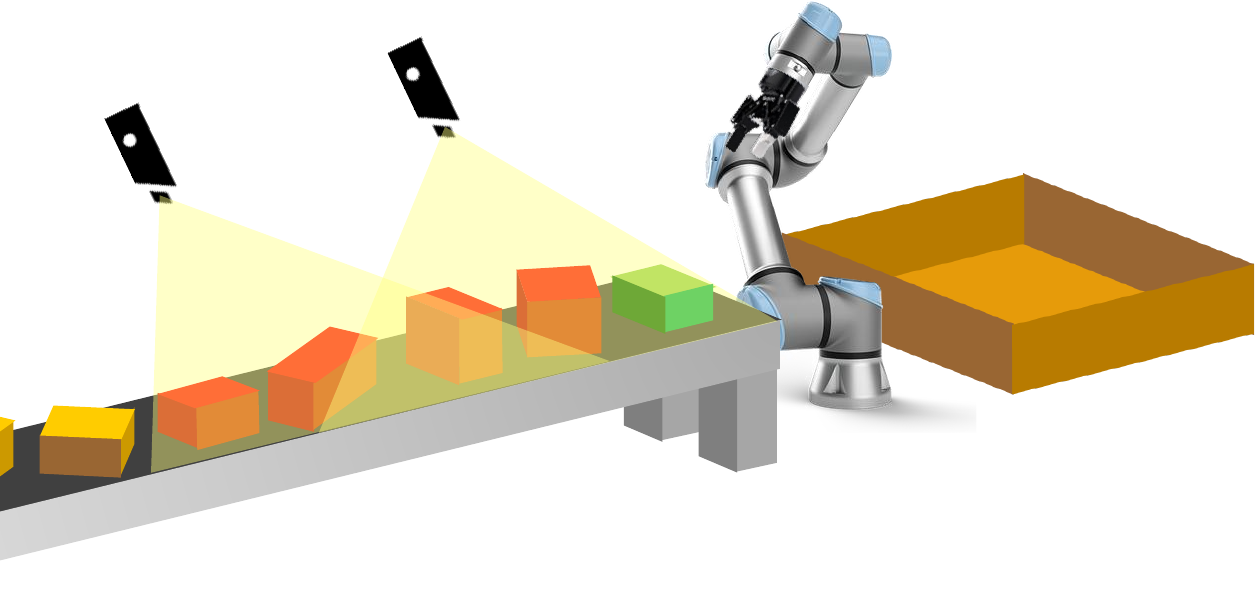}
        
        {\fontsize{8pt}{10pt}\selectfont (b)}
    \end{minipage}%
    \hfill
    \begin{minipage}{0.49\columnwidth}
        \centering
        \includegraphics[width=1.0\linewidth]{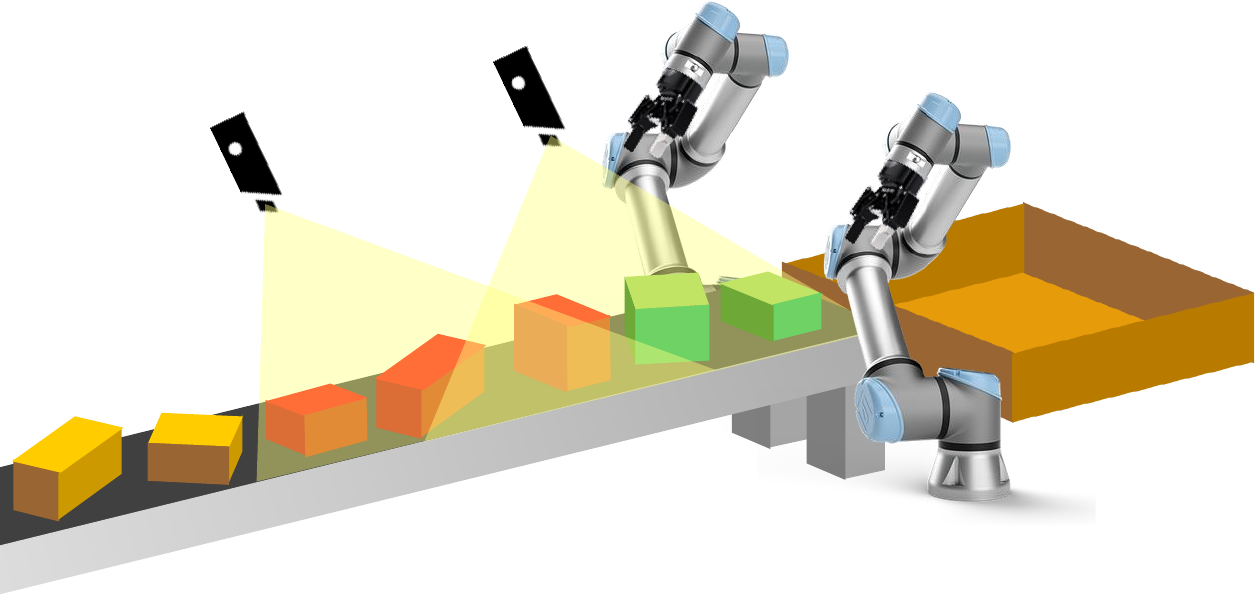}
        
        {\fontsize{8pt}{10pt}\selectfont (e)}
    \end{minipage}

    \par\medskip

    \begin{minipage}{0.49\columnwidth}
        \centering
        \includegraphics[width=1.0\linewidth]{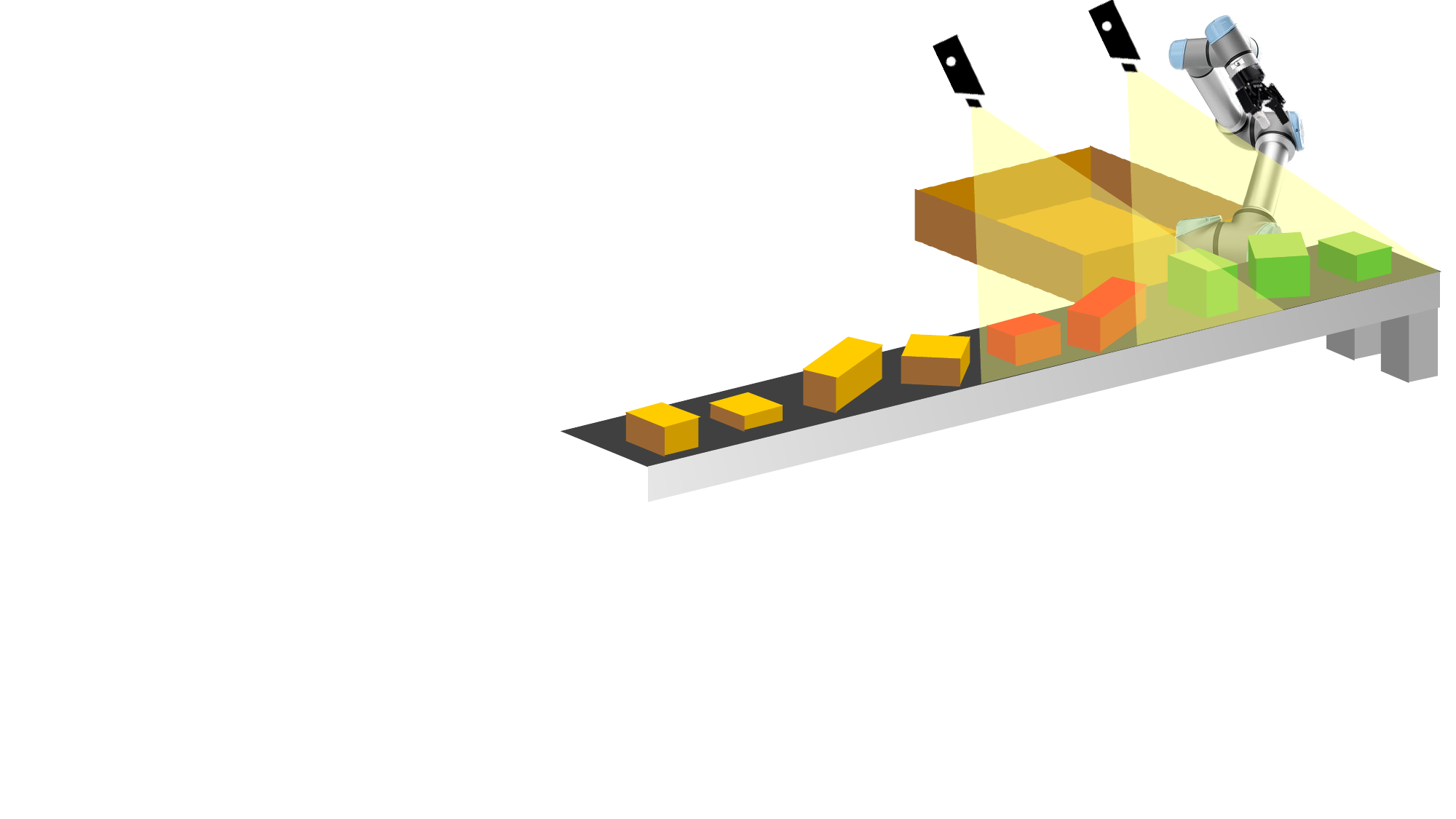}
        
        {\fontsize{8pt}{10pt}\selectfont (c)}
    \end{minipage}%
    \hfill
    \begin{minipage}{0.49\columnwidth}
        \centering
        \includegraphics[width=1.0\linewidth]{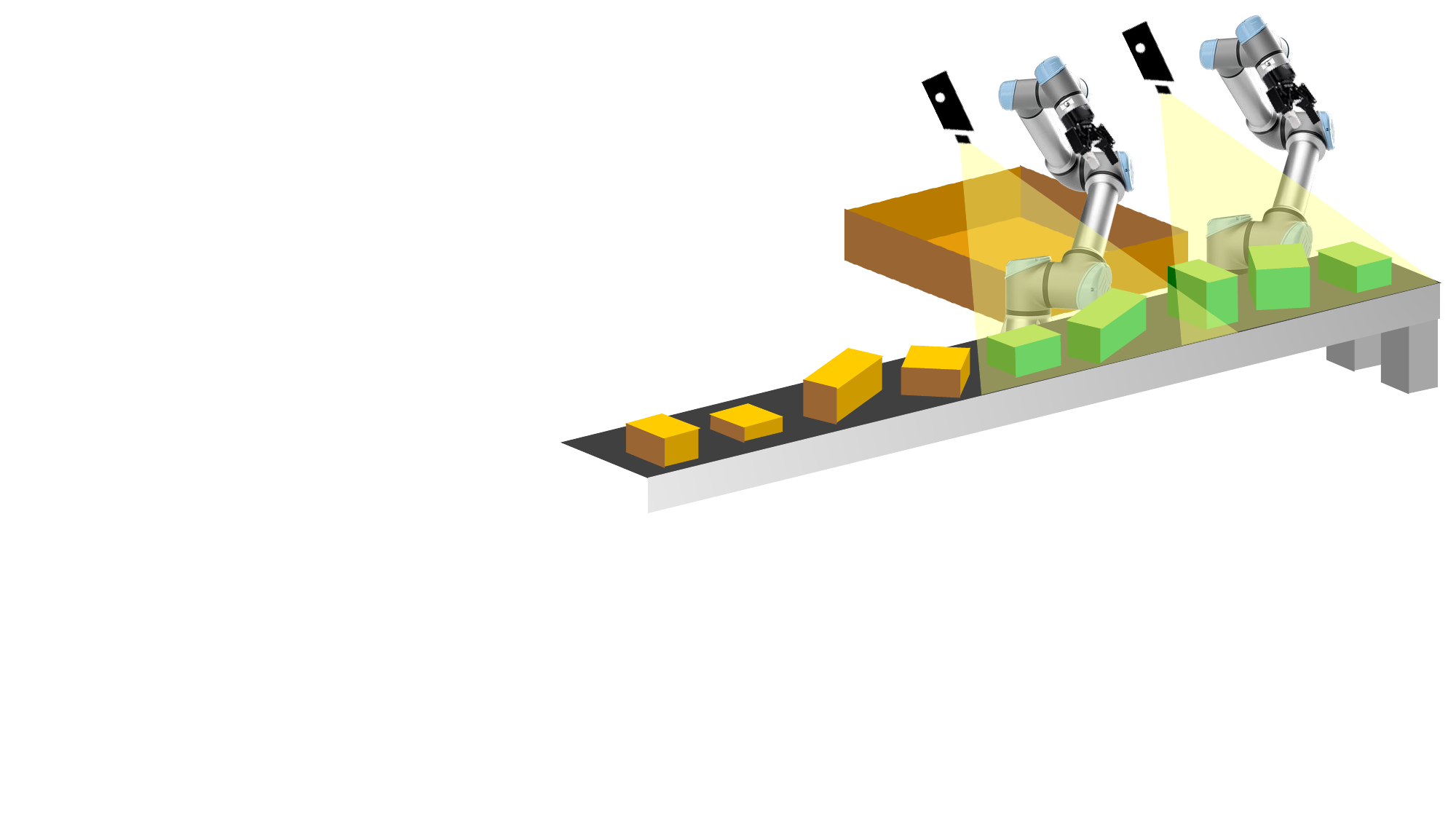}
        
        {\fontsize{8pt}{10pt}\selectfont (f)}
    \end{minipage}

    \caption{Bin packing systems in six different scenarios.
    (a--c) illustrate configurations with a single manipulator, whereas (d--f) depict those involving dual manipulators.
    The number of known items (in green and red) varies across scenarios, and the number and placement of manipulators determine which items are accessible.}
    \label{fig:scenarios}
\end{figure}

BPP includes offline, online, and semi-online variants, depending on the availability of item information~\cite{semi-online}. While the offline BPP assumes full knowledge of all items, real-world applications typically involve online or semi-online settings, where the information of items are revealed sequentially. Semi-online BPPs are more common in industrial contexts such as logistics and manufacturing.
In typical warehouse or inspection setups, a camera or barcode reader is installed above a conveyor belt, allowing partial observation of incoming items within a fixed field of view. In such cases, the system can access limited item information (e.g., size, orientation, or label data) before packing, which aligns with the assumptions of the 2D online and semi-online settings. % review 1-2
In such settings, strategies for managing uncertainty, such as repacking, can improve adaptability to changes in item arrivals. BPP is also categorized by packing dimension (e.g., 2D, 3D)~\cite{multidimensionalReview}.
% Although there has been a line of research on both 2D BPP and 3D BPP, 2D BPP remains underexplored, particularly in achieving near-optimal packing through rotation and/or repacking.
% review 1-1
Although there has been a line of research on both 2D BPP and 3D BPP, many real-world applications are effectively 2D due to physical or operational constraints. 
For example, fragile or breakable items such as glass panels and ceramics cannot be stacked; 
inspection or vision-based operations often require items to remain within a single camera view; 
and layout optimization on flat materials such as textiles, leather, or printed circuit boards (PCBs) is inherently 2D. 
In addition, certain industrial containers, such as interlocking plastic crates (e.g., milk crates), and uniformly shaped items that can be stacked in shallow multi-layer configurations, also fit naturally within a 2D formulation. 
In these environments, the packing task is dominated by footprint efficiency rather than vertical stacking. 
Motivated by these practical use cases, this work focuses on the 2D online and semi-online bin packing problem.

Most existing BPP methods assume a single-robot setting, although coordinating multiple manipulators can significantly reduce execution time through parallelization. Coordinating multiple robots for BPP remains computationally challenging due to the inherent complexity of multi-robot task allocation and multi-arm motion planning~\cite{stopngo, SynchronizedDualArm, multi-agentGame, ahn2022coordination}.
While asynchronous or semi-synchronous operation can potentially improve execution efficiency, these approaches considerably increase the complexity and computation time of task and motion planning required to ensure safe and collision-free coordination. Therefore, the proposed framework adopts a synchronous operation, which provides predictable coordination between manipulators, simplifies collision avoidance, and reduces planning complexity, which, makes it more practical for real-world deployment.

This paper addresses the 2D bin packing problem (BPP) under practical scenarios involving varying numbers of known items, accessible items, and manipulators, as illustrated in Fig.~\ref{fig:scenarios}.  
To handle such diverse configurations and support repacking for improved bin utilization, we develop a hierarchical approach that integrates deep reinforcement learning (DRL) with heuristic search.  
The low-level DRL agent, trained with Asynchronous Advantage Actor-Critic (A3C)~\cite{A3C}, selects item positions based on a given bin and item state.  
On top of this, the high-level search constructs a tree by virtually packing items at positions proposed by the low-level agent, exploring different packing orders and orientations.  
Unpacking and repacking can be considered to improve bin utilization.  
A best task sequence is then selected from the candidate branches in the tree and translated into executable actions through integrated task and motion planning, ensuring physical feasibility in both single- and dual-arm settings.  
In dual-arm scenarios, items are assigned to manipulators based on accessibility, and coordinated actions are planned for parallel execution using off-the-shelf motion planners.

To the best of our knowledge, this is the first work to solve the 2D BPP using a dual-manipulator system under diverse industrial conditions described in Fig.~\ref{fig:scenarios}. The main contributions of this paper are:

\begin{itemize}
\item A DRL framework that outperforms the state-of-the-art 2D online BPP methods with significantly higher bin utilization.
\item A heuristic search algorithm that integrates robot accessibility and multi-manipulator coordination into the policy generation process, enabling feasible and optimized packing decisions under offline, semi-online, and online conditions with varying perception ranges.
\item A repacking strategy that consistently achieves near-100\% bin utilization.
\end{itemize}

\section{Related Works}

Prior work has addressed offline and online 2D BPPs, with recent advances extending to semi-online settings and incorporating rotation and repacking techniques. This section reviews related research on these BPP variants and strategies for rotation and repacking. We also discuss approaches for coordinating multi-robot operations in shared workspaces. Table~\ref{tab:dimensional_comparison} summarizes representative 2D and 3D BPP studies, focusing on their problem settings and robotic implementation aspects such as repacking, multi-bin handling, and dual-arm coordination.

% \begin{table}[t]
% \centering
% \caption{Comparison of representative studies in 2D and 3D bin packing settings.}
% \label{tab:dimensional_comparison}
% \begin{tabular}{lcc}
% \toprule
% \textbf{Characteristic} & \textbf{2D} & \textbf{3D} \\
% \midrule
% Offline / Semi-online                & \textbf{Ours}, \cite{genetic2D} & ~\cite{3Dheuristic,Metaheuristic,solving,aaaik-BPP,feasible3DBPP,3Dvision,Synergies,Heuritics_integratedDRL3D,dual_conveyor,dual_arm_springer}    \\
% Online                               & \textbf{Ours},\cite{deeppack,brain-inspired,Skyline,ShelfNextFit}    & \cite{aaaik-BPP,feasible3DBPP,3Dvision,Synergies,Heuritics_integratedDRL3D}    \\
% Repacking                            & \textbf{Ours}                                              & \cite{Synergies,Heuritics_integratedDRL3D,dual_conveyor}    \\
% Multi-bin                             &                                                      & \cite{3Dheuristic,Metaheuristic,aaaik-BPP,dual_conveyor}    \\
% Robotic implementation               & \textbf{Ours}                                                & \cite{Metaheuristic,aaaik-BPP,feasible3DBPP,Synergies,Heuritics_integratedDRL3D,dual_conveyor,dual_arm_springer}    \\
% Dual-arm per bin  & \textbf{Ours}                                                \\
% \bottomrule
% \end{tabular}
% \end{table}

\begin{table}[t]
\centering
\caption{Comparison of representative studies in 2D and 3D bin packing settings.}
\label{tab:dimensional_comparison}
\scalebox{0.90}{
\begin{tabular}{lcc}
\toprule
\textbf{Characteristic} & \textbf{2D} & \textbf{3D} \\
\midrule
Offline / Semi-online                & \textbf{Ours}, \cite{genetic2D}                                       & \cite{3Dheuristic,Metaheuristic,solving,aaaik-BPP,feasible3DBPP,3Dvision,Synergies,Heuritics_integratedDRL3D,dual_conveyor,dual_arm_springer}    \\
Online                               & \textbf{Ours},\cite{deeppack,brain-inspired,Skyline,ShelfNextFit}    & \cite{aaaik-BPP,feasible3DBPP,3Dvision,Synergies,Heuritics_integratedDRL3D}    \\
Repacking                            & \textbf{Ours}                                              & \cite{Synergies,Heuritics_integratedDRL3D,dual_conveyor}    \\
Multi-bin                             &                                                      & \cite{3Dheuristic,Metaheuristic,aaaik-BPP,dual_conveyor}    \\
Robotic Implementation               & \textbf{Ours}                                                & \cite{Metaheuristic,aaaik-BPP,feasible3DBPP,Synergies,Heuritics_integratedDRL3D,dual_conveyor,dual_arm_springer}    \\
Dual-arm Allocation per bin  & \textbf{Ours}                                                \\
\bottomrule
\end{tabular}
}
\end{table}

In the offline BPP where all items are known, metaheuristic approaches such as genetic algorithms~\cite{genetic2D}, Tabu Search~\cite{3Dheuristic}, and layer-building strategies~\cite{Metaheuristic} have been proposed. A hybrid DRL-heuristic method~\cite{solving} also aims to minimize item surface area. However, these methods often suffer from limited generalizability~\cite{genetic2D}, sensitivity to initialization and local optima~\cite{3Dheuristic}, reliance on handcrafted rules~\cite{Metaheuristic}, or lack of interaction between learning and rule-based components~\cite{solving}.

In online BPP where items arrive sequentially, heuristic methods such as Skyline~\cite{Skyline} and ShelfNextFit~\cite{ShelfNextFit} have been proposed. Some account for item orientations~\cite{generalized3D} but generalize poorly across varying item sizes. DRL-based methods improve generalization~\cite{deeppack, brain-inspired, Recent3DRearrangement} but cannot revise past decisions, limiting near-optimal bin utilization.

The semi-online BPP, where some future items are partially known, has drawn attention as it reflects realistic industrial scenarios. While packing the current accessible item, partial knowledge of upcoming items is available. In~\cite{lookahead}, a bounded lookahead heuristic estimates processing costs based on known future items. BPP-$k$~\cite{aaaik-BPP} extends an RL model trained on BPP-1 to look $k$ steps ahead via tree search. Other works~\cite{feasible3DBPP,3Dvision} also employ this approach. However, these approaches do not consider robot accessibility, leaving executability unverified. Moreover, none consider repacking placed items.

To achieve high bin utilization, repacking has emerged as a crucial strategy. \cite{Heuritics_integratedDRL3D} integrates heuristic rules and DRL: a heuristic determines whether to pack or unpack based on wasted volume, and the RL agent selects the position. \cite{Synergies} compares values from dual critic networks to decide between packing and unpacking. However, both~\cite{Heuritics_integratedDRL3D, Synergies} are limited by myopic item selection (e.g., random or first-in-last-out) without global optimization.
Additionally,~\cite{Synergies} compares unpacked and packed states unfairly, since the unpacked state has fewer packed items and thus an advantage in value.

Recent studies have also explored multi-arm in bin packing scenarios.
~\cite{dual_conveyor} proposes a deep reinforcement learning framework for online and concurrent 3D bin packing with bin replacement, in which multiple bins are processed in parallel to improve throughput.
Their work focuses on optimizing packing efficiency across multiple containers, rather than coordinating multiple manipulators within a single bin.
On the other hand, ~\cite{dual_arm_springer} develops a dual-arm robotic packing framework emphasizing joint motion coordination and trajectory planning for cooperative manipulation of a single object.
While both studies address related challenges in robotic packing, their problem formulations differ from ours: we focus on dual-manipulator cooperation for packing multiple objects within a single bin under online and semi-online settings, where task allocation and placement decisions are jointly optimized through a hierarchical algorithm. % review 1-5

While multi-robot manipulation improves packing throughput, it introduces coordination challenges. Some works have addressed item rearrangement in shared workspaces~\cite{SynchronizedDualArm, multi-agentGame, ahn2022coordination}, but coordinated bin packing remains largely unexplored.

% To fill this gap, we propose a framework for 2D BPP under realistic multi-robot settings.
% Our method considers item accessibility through forward simulation to ensure feasibility, and integrates coordinated task planning and repacking within a unified hierarchical structure. Leveraging a dual-manipulator system, we enable parallel execution while ensuring packing decisions are both feasible and optimized for bin utilization.
% To fill this gap, we propose a framework for 2D BPP under realistic multi-robot settings. 
% Unlike prior studies that primarily focused on single-arm, our method explicitly considers dual-manipulator coordination within a single bin. 
% The proposed approach incorporates a forward simulation-based feasibility validation that evaluates accessibility before placement decisions, 
% integrates synchronized dual-arm task allocation and workspace-aware motion planning into a unified hierarchical structure, 
% and adopts a repacking mechanism that enhances utilization and stability. 
% By combining these elements, our framework achieves both realistic feasibility reasoning and near-optimal bin utilization, distinguishing it from previous studies.

To fill this gap, we propose a framework for 2D BPP under realistic dual-arm settings.
Unlike previous studies that determine object placement and orientation before considering robotic execution, our method jointly plans the packing policy while accounting for the reachability of both single and dual manipulators.
This integrated formulation ensures that packing decisions are physically feasible for robot execution while maintaining near-optimal packing efficiency.
Moreover, the framework synergistically integrates a low-level RL policy with a high-level tree search, combining the flexibility of learning-based decision making with the reliability of search-based reasoning. % review 1-6

\section{Problem Formulation}

\begin{figure*}[t]
    \centering

    \begin{minipage}[b]{0.37\textwidth}
        \centering
        \includegraphics[width=0.9\linewidth]{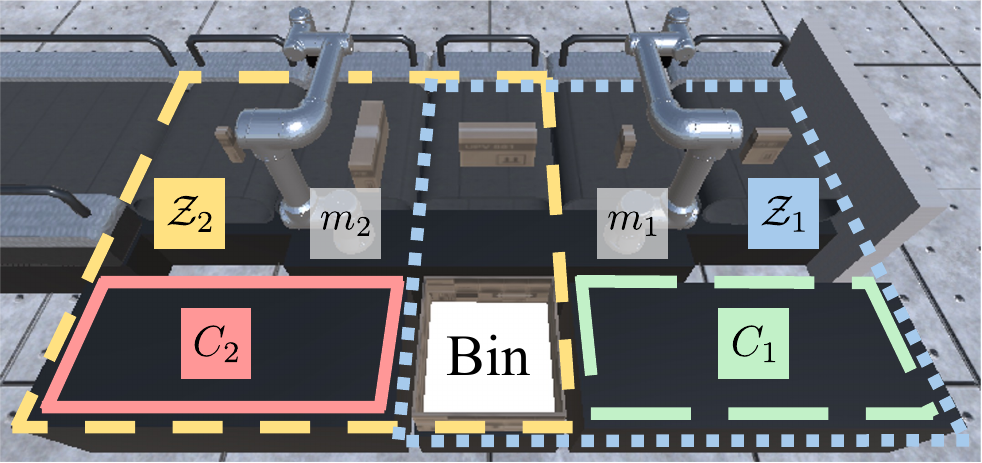}
        % \vspace{2pt}
        
        {\fontsize{8pt}{10pt}\selectfont (a) The task space and cache of the dual manipulators}
        \label{fig:workspace}
    \end{minipage}
    % \hspace{2pt}
    \begin{minipage}[b]{0.28\textwidth}
        \centering
        \includegraphics[width=1.0\linewidth]{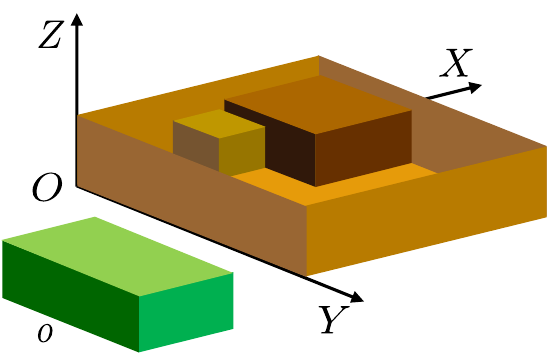}
        % \vspace{2pt}

        {\fontsize{8pt}{10pt}\selectfont (b) The bin and an item}
        \label{fig:cuboid}
    \end{minipage}
    % \hspace{2pt}
    \begin{minipage}[b]{0.33\textwidth}
        \centering
        \includegraphics[width=1.0\linewidth]{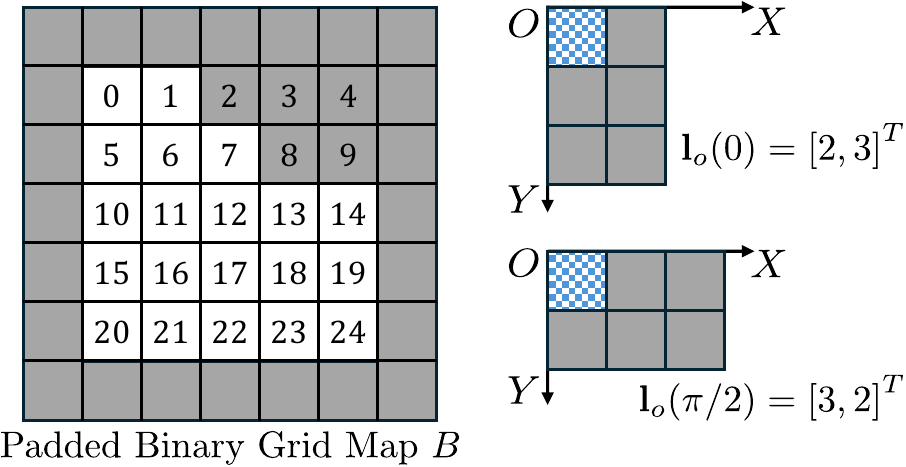}
        % \vspace{2pt}

        {\fontsize{8pt}{10pt}\selectfont (c) The representation of the bin and the item}
        \label{fig:matrix}
    \end{minipage}

    \caption{Illustration of the Bin Packing system.
    (a) The manipulator is denoted by $m$, $\mathcal{Z}$ represents the task space of the manipulator, and $C$ is the temporary storage area for unpacked items.
    (b) The bin is an open cuboid, and the next item $o$ is a solid cuboid.
    (c) The bin is modeled as a padded binary grid map $B$, where each cell corresponds to a low-level position action indexed by its coordinate in the image coordinate system. The values increase row-wise from left to right and top to bottom, with the origin located at the top-left corner. The item is encoded as the rotated vector $\mathbf{l}(\phi)$, and a checkerboard-patterned pixel indicates the top-left corner of the placed item.
}
    \label{fig:env}
\end{figure*}

Our 2D BPP is formulated as a hierarchical Markov Decision Process (MDP). The high-level MDP focuses on finding a sequence of task primitives determining what to pack or unpack, based on the current bin and items. The low-level MDP determines the precise positions to execute these primitives. After each high-level action is executed, the system transitions to the next state with an updated bin configuration.

As shown in Figs.~\ref{fig:env}(a) and~\ref{fig:env}(b), we consider a 2D BPP environment where the bin is modeled as an open cuboid and items as closed cuboids, all projected onto the $xy$-plane. We assume up to two robotic manipulators, denoted by $m_i$ for $i \in \{1,2\}$, each with a task space $\mathcal{Z}_i \subset \mathcal{W} \subseteq \mathbb{R}^3$. Items arrive sequentially via a conveyor belt (or other similar facilities), which advances as items are removed.
The bin and each item $o$ are discretized into grid cells: the bin is a $W \times H$ grid, and item $o$ occupies $w_o \times h_o$ cells, where $W, H, w_o, h_o \in \mathbb{Z}^+$. The bin occupancy is represented by a padded binary map $B \in \{0,1\}^{(W+2)\times(H+2)}$, where $0$ and $1$ indicate free and occupied cells. The outermost rows and columns are padded with $1$s to represent the bin boundaries (see Fig.~\ref{fig:env}(c)).
The rotated size vector $\mathbf{l}_o(\phi) = [l_x^{(o,\phi)}, l_y^{(o,\phi)}]^T$ defines the dimensions of item $o$ after rotation by $\phi \in \{0, \frac{\pi}{2}\}$. For example, if $w_o = 3$ and $h_o = 1$, then $\mathbf{l}_o(0) = [3,1]^T$ and $\mathbf{l}_o(\frac{\pi}{2}) = [1,3]^T$.
We denote $\mathcal{N}$ as items on the conveyor, $\mathcal{C}$ as unpacked items, and $\mathcal{I}$ as packed items. Each manipulator $m_i$ has a buffer space $C_i \subset \mathcal{Z}_i$ to temporarily store items in $\mathcal{C}$.

% As shown in Figs.\ref{fig:env}(a)--\ref{fig:env}(b), we consider a 2D BPP environment where the bin is modeled as an open cuboid and items as closed cuboids projected onto the $xy$-plane.
% Up to two manipulators $m_i$ ($i \in {1,2}$) operate within their workspaces $\mathcal{Z}_i \subset \mathcal{W} \subseteq \mathbb{R}^3$. Items arrive sequentially via a conveyor belt, which advances as items are removed.
% The bin is a $W \times H$ grid, and each item $o$ occupies $w_o \times h_o$ cells, with $W, H, w_o, h_o \in \mathbb{Z}^+$. The bin occupancy is encoded as a padded binary matrix $B \in {0,1}^{(W+2)\times(H+2)}$, where outer padding with $1$s represents bin boundaries (Fig.\ref{fig:env}(c)).
% The rotated size vector $\mathbf{l}_o(\phi) = [l_x^{(o,\phi)}, l_y^{(o,\phi)}]^T$ specifies item dimensions under rotation $\phi \in {0, \frac{\pi}{2}}$; e.g., $w_o=3$, $h_o=1$ yields $\mathbf{l}_o(0) = [3,1]^T$ and $\mathbf{l}_o(\frac{\pi}{2}) = [1,3]^T$.
% We define $\mathcal{N}$ as items on the conveyor, $\mathcal{C}$ as unpacked items, and $\mathcal{I}$ as packed items. Each manipulator has a buffer $C_i \subset \mathcal{Z}_i$ for storing $\mathcal{C}$.

\subsection{The High-level MDP}

The high-level MDP governs overall task planning, which is defined as $(\mathcal{S}_{\text{high}}, \mathcal{A}_{\text{high}}, P_{\text{high}}, R_{\text{high}})$.
Each state $s_{\text{high}} \in \mathcal{S}_{\text{high}}$ comprises the bin occupancy $B$, the set of items to be packed $\mathcal{N}$, the packed items $\mathcal{I}$, and the unpacked items $\mathcal{C}$:
\[
s_{\text{high}} = \{B, \mathcal{N}, \mathcal{I}, \mathcal{C}\}.
\]

An action $a_{\text{high}} \in \mathcal{A}_{\text{high}}$ is a sequence of task primitives: $\texttt{pack}$, $\texttt{unpack}$, or $\texttt{terminate}$. The $\texttt{pack}$ and $\texttt{unpack}$ primitives have parameters which depend on specific items, while $\texttt{terminate}$ ends the current episode when no further action is necessary or feasible.

The primitive $\texttt{pack}(o, \phi_o, x_o, y_o)$ places item $o$ from $\mathcal{N}$ or $\mathcal{C}$ into the bin with its top-left corner at $(x_o, y_o)$ and orientation $\phi_o$. While $\phi_o$ is selected by the high-level MDP, $(x_o, y_o)$ is determined by the low-level MDP. This operation updates the bin occupancy by setting the occupied cells to $1$:
\begin{equation}
B^{\prime}(x, y) = 1
\label{eq:update}
\end{equation}
for all $x \in [x_o+1, x_o + l_x^{(o,\phi_o)}]$ and $y \in [y_o+1, y_o + l_y^{(o,\phi_o)}]$. The item $o$ is then removed from $\mathcal{N}$ or $\mathcal{C}$ and added to $\mathcal{I}$.

The primitive $\texttt{unpack}(o)$ removes item $o$ from the bin, setting its previously occupied cells to:
\[
B^{\prime}(x, y) = 0
\]
for all $x \in [{px}_{o} + 1, {px}_{o} + l_x^{(o,{\psi_{o}})}]$ and $y \in [{py}_{o} + 1, {py}_{o} + l_y^{(o,\psi_{o})}]$,
where $({px}_{o}, {py}_{o})$ and $\psi_{o}$ denote the position and orientation of $o$ before unpacking. The item is removed from $\mathcal{I}$ and added to $\mathcal{C}$.

A complete action is composed by sequencing these primitives. For example:
\[
\begin{aligned}
a_{\text{high}} = (&\texttt{unpack}(o_1), \texttt{unpack}(o_2), \\
&\texttt{pack}(o_1, \phi_{o_1}, x_{o_1}, y_{o_1}), \texttt{pack}(o_2, \phi_{o_2}, x_{o_2}, y_{o_2}), \\
&\texttt{terminate})
\end{aligned}
\]
unpacks $o_1$ and $o_2$, then repacks them at new positions, followed by termination. An action may consist solely of \texttt{terminate}, or include only \texttt{pack}.

The high-level reward \(r_{\text{high}}\) is defined as the total change in bin occupancy resulting from the executed primitives in \(a_{\text{high}}\):
\[
r_{\text{high}}(s_\text{high}, a_\text{high}) = \sum_{\lambda \in a_{\text{high}}} c_o w_o h_o,
\]
where \(c_o = 1\) if \(\lambda\) is \texttt{pack}, \(c_o = -1\) if \(\lambda\) is \texttt{unpack}, and \(c_o = 0\) otherwise.
The cumulative high-level reward indicates the bin utilization ratio, encouraging space-efficient packing in real-world applications.

\subsection{The Low-level MDP}

As mentioned, the primitive $\texttt{pack}(o, \phi_o, x_o, y_o)$ requires a specific position $(x_o, y_o)$ to place item $o$ in the bin. When constructing $a_{\text{high}}$, the low-level MDP \((\mathcal{S}_{\text{low}}, \mathcal{A}_{\text{low}}, P_{\text{low}}, R_{\text{low}}, \gamma_{\text{low}})\) is used to determine this position. The low-level state for item $o$ under orientation $\phi \in \{0, \frac{\pi}{2}\}$ is defined as:
\[
s_{\text{low}}^{(o,\phi)} = \{B, \mathbf{l}_o(\phi)\},
\]
where $B$ is the current bin occupancy and $\mathbf{l}_o(\phi)$ is the rotated dimension of $o$. Given $s_{\text{low}}^{(o,\phi)}$, the low-level action \(a_{\text{low}}^{(o,\phi)} \in \mathcal{A}_{\text{low}}\) specifies the position for placing the top-left corner of $o$ in the bin. The high-level MDP evaluates both orientations and selects one along with its corresponding position to complete the \texttt{pack} primitive.
The bin position is represented by an index from $0$ (top-left) to $W \cdot H - 1$ (buttom-right), following a row-wise order as illustrated in Fig.~\ref{fig:env}(c).
Given a position $(x_o, y_o)$ for item $o$, the low-level action is defined as:
\[
a_{\text{low}}^{(o,\phi)} = x_o + y_o \cdot W.
\]
If no valid position is available, a special \textit{no-position} action is selected:
\begin{equation}
    a_{\text{low}}^{(o,\phi)} = W \cdot H. \label{eq:nop}
\end{equation}
Thus, the low-level action space is defined as $\mathcal{A}_{\text{low}} = \{0, 1, \dots, W \cdot H\}$. For each $\phi \in \{0, \frac{\pi}{2}\}$, $a_{\text{low}}^{(o,\phi)}$ is chosen in state $s_{\text{low}}^{(o,\phi)}$ to place item $o$ of dimension $l_x^{(o,\phi)} \times l_y^{(o,\phi)}$.

The low-level reward is defined as the number of occupied cells adjacent to the packed item $o$:

\begin{flalign}
& r_{\text{low}}(s_{\text{low}}^{(o,\phi)}, a_{\text{low}}^{(o,\phi)}) = \notag\\
& + \sum_{j=1}^{l_x^{(o,\phi)}} \left( B(x_o + j, y_o) + B(x_o + j, y_o + l_y^{(o,\phi)}) \right) \notag\\
& + \sum_{j=1}^{l_y^{(o,\phi)}} \left( B(x_o, y_o + j) + B(x_o + l_x^{(o,\phi)}, y_o + j) \right).
\label{eq:r_low}
\end{flalign}
A higher reward is given when $o$ is placed adjacent to more occupied cells; zero reward is given otherwise. The low-level MDP aims to maximize the expected cumulative reward:
\[
\sum_{j=0}^{\infty} \gamma_{\text{low}}^j \, r_{\text{low}}(s_{\text{low}}^{(o_j,\phi)}, a_{\text{low}}^{(o_j,\phi)}),
\]
where the discount factor $\gamma_{\text{low}}$ controls the contribution of future rewards. This objective encourages policies that promote compact placements to improve long-term bin utilization.

\section{Methods}

\begin{figure}[!t]
    % \centering
    
    \centerline{\includegraphics[width=\columnwidth]{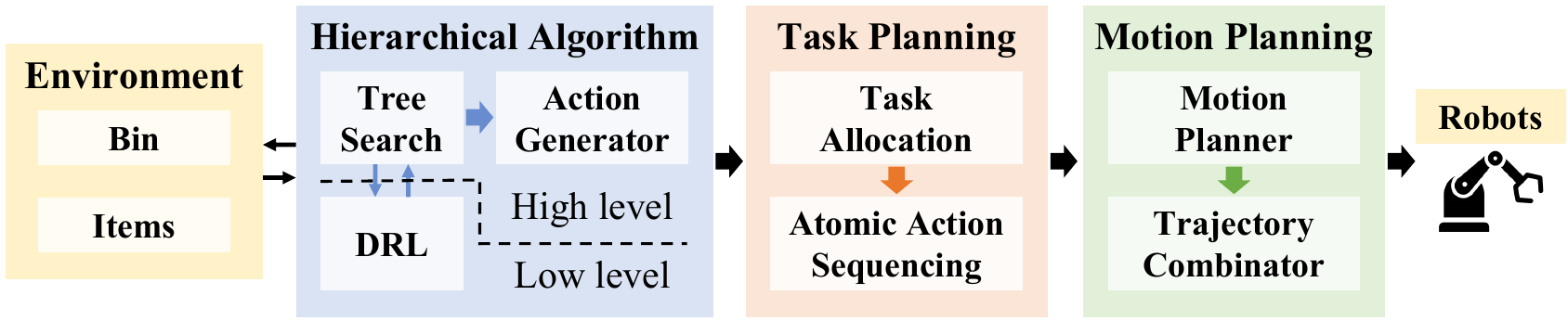}}
    \caption{Overall framework. Our bin packing system integrates a hierarchical algorithm with task and motion planning.}
    \label{fig:framework}
    % \vspace{-10pt}
\end{figure}

The overall framework, illustrated in Fig.~\ref{fig:framework}, consists of three main modules: a hierarchical algorithm, task planning, and motion planning.

To solve the 2D BPP, we propose a hierarchical algorithm that integrates DRL with heuristic search (Fig.~\ref{fig:hierarchical}). This hierarchical structure enables interaction between the high- and low-level MDPs through a tree search process.
At the low level (Sec.~\ref{sec:low}), an RL agent generates \textit{positional actions} that determine precise placement positions within the bin. At the high level (Sec.~\ref{sec:high}), the controller operates in two stages. In the first stage, a \textit{Depth-First Selective Beam Search (DFS-BS)} algorithm expands a search tree by recursively packing items at positions proposed by the low-level agent, while exploring various packing orders and orientations. DFS-BS controls the branching factor by selecting the most promising candidates based on reward values and item sizes. When necessary, the search performs unpacking operations and considers repacking to improve bin utilization.
In the second stage, each candidate packing sequence (i.e., each branch of the tree) is evaluated via forward simulation using a reward-based heuristic score, and the highest-scoring sequence is selected as the high-level action. Each high-level action either specifies a set of items to unpack or determines a set of items to pack, including their placement positions and orientations. Executing a high-level action transitions the environment to a new state with updated bin occupancy and item sets.

Subsequently, the task planning module allocates the task primitives from the resulting high-level actions to individual robots and translates them into atomic actions (e.g., \textit{pick} and \textit{place}). The motion planning module then generates joint trajectories for each robot to execute the atomic actions in a physically feasible manner.

\subsection{Low-Level: Position Selection}
\label{sec:low}

\begin{figure}[!t]
    
\centerline{\includegraphics[width=\columnwidth]{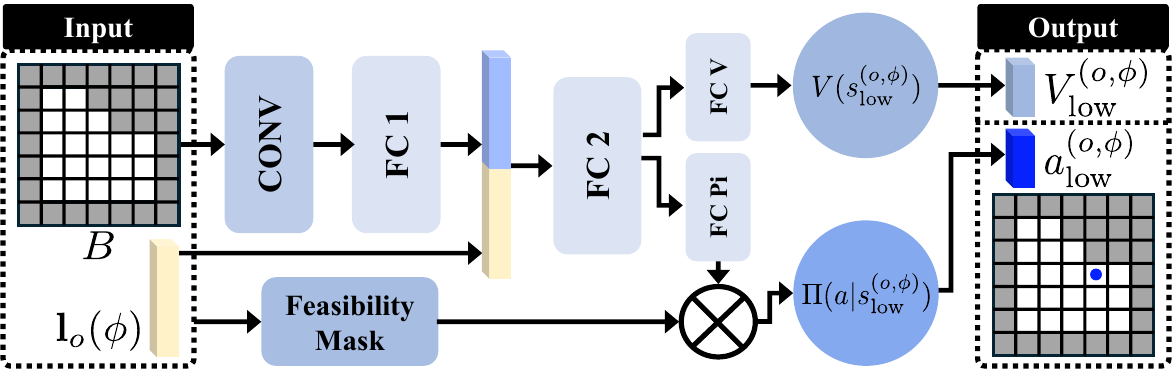}}
    \caption{Actor-critic framework. The input consists of the current bin configuration $B$ and the size vector $\mathbf{l}(\phi)$ of the item.}
    \label{fig:drl}
    % \vspace{-10pt}
\end{figure}

The low-level MDP is solved by an RL agent that determines the packing position for each item. The agent employs A3C, an actor-critic algorithm, to learn the optimal policy $\pi_\text{low}$, which generates $a_{\text{low}}^{(o,\phi)}$ in $s_{\text{low}}^{(o,\phi)}$ for item $o$ with candidate orientation $\phi$, as defined in~\eqref{eq:policy}. A3C enables asynchronous training across multiple environments and leverages a value function to evaluate states for improved decision-making. The actor-critic framework is illustrated in Fig.~\ref{fig:drl}.

\subsubsection{State input}

The agent takes $B$ and $\mathbf{l}_o(\phi)$ as input. A convolutional neural network (CNN) encodes $B$ into a feature vector, concatenated with $\mathbf{l}_o(\phi)$ and passed through a linear layer. The actor outputs action probabilities $\Pi(a \mid s^{(o,\phi)}_\text{low})$, and the critic estimates the state value $V(s^{(o,\phi)}_\text{low})$.

\subsubsection{Feasibility mask}

The placement of item $o$ must consider both the bin boundary and existing packed items. While $o$ should occupy free grid cells within the bin, it is also important to avoid selecting the no-position action when sufficient space exists. To enforce this, we compute a \textit{feasibility mask} $\mathbf{b}$ at each state before action selection:
$
\mathbf{b} = [b_0, b_1, \ldots, b_{W \cdot H}],
$
where $b_j = 0$ if action $a_{\text{low}} = j$ violates placement constraints, and $b_j = 1$ otherwise.
Inspired by~\cite{Synergies}, we apply the mask $\mathbf{b}$ to the pre-softmax logits $\mathbf{z} = [z_0, z_1, \ldots, z_{W \cdot H}]$ from the linear layer:
$
\mathbf{z}' = \mathbf{b} \odot \mathbf{z} + (1 - \mathbf{b}) \odot (-10^8),
$
where $\odot$ denotes element-wise multiplication. This effectively suppresses actions that violate placement constraints, including the no-position action when sufficient space is available, by assigning large negative scores. The final policy is defined via softmax:
\begin{equation}
    \pi(a_t \mid s_t) = \text{softmax}(\mathbf{z}').
\end{equation}

\subsubsection{Loss function}

The A3C algorithm employs parallel actor-learners with a shared global network. Each actor updates the policy $\pi(a_t|s_t; \theta)$ and value function $V(s_t; \theta_v)$ using local rollouts. The policy is trained via the policy gradient:
\[
\nabla_\theta J(\theta) = \mathbb{E}_{\pi_\theta} \left[ \nabla_\theta \log \pi(a_t|s_t; \theta) A(s_t, a_t) \right],
\]
where the advantage $A(s_t, a_t) = Q(s_t, a_t) - V(s_t)$. The actor and critic losses are defined as:
\[
L^{\text{actor}} = -\log \pi(a_t|s_t; \theta) A(s_t, a_t),
\]
\[
L^{\text{critic}} = \left( R_t + \gamma V(s_{t+1}) - V(s_t) \right)^2.
\]
The total loss $L_{\text{total}}$ is the sum of both. Gradients are asynchronously aggregated to update $\theta$ and $\theta_v$.

\subsection{Interaction Between High-level and Low-level}

We introduce a tree $\mathcal{T} = (\mathcal{V}, \mathcal{E})$ that unifies the high-level and low-level MDPs, where $\mathcal{V}$ and $\mathcal{E}$ denote the sets of vertices and edges, respectively. The tree is used at the high level to evaluate permutations of task primitives along with their parameters. For each high-level state $s_{\text{high}} = \{B, \mathcal{N}, \mathcal{I}, \mathcal{C}\}$, a tree $\mathcal{T}$ is constructed, where each vertex $v \in \mathcal{V}$ represents a high-level state of the same form.
Using the low-level policy $\pi_{\text{low}}$, a low-level action $a_{\text{low}}^{(o,\phi)}$ is generated for item $o$ with orientation $\phi$, producing a child vertex:
\begin{equation}
a_{\text{low}}^{(o,\phi)} = \pi_{\text{low}}(s_{\text{low}}^{(o,\phi)}) = \pi_{\text{low}}(\{B, \mathbf{l}_o(\phi)\}). \label{eq:policy}
\end{equation}
This leads to a new vertex $v' = \{B', \mathcal{N}', \mathcal{I}', \mathcal{C}'\} \in \mathcal{V}$ in the tree, where $B'$ is updated via \eqref{eq:update}, and item $o$ is moved from $\mathcal{N}$ or $\mathcal{C}$ to $\mathcal{I}$.\footnote{Item $o$ is removed from $\mathcal{N}$ or $\mathcal{C}$, whichever contains it, and added to $\mathcal{I}$.}
If $a_{\text{low}}^{(o,\phi)}$ is the no-position action \eqref{eq:nop}, no child is generated and $v$ becomes a leaf vertex. A leaf represents a state where either (i) no items remain in $\mathcal{N} \cup \mathcal{C}$, or (ii) $o$ cannot be placed due to collisions with packed items in $\mathcal{I}$.

From a sequence of vertices from the root of $\mathcal{T}$ and to a leaf, we extract a tuple $(o, \phi_o, a_{\text{low}}^{(o,\phi)}, d(v'))$ for each child $v'$ of $v$, where $d(v') = d(v) + 1$ denotes the depth of $v'$ in $\mathcal{T}$. Excluding the root, we construct a sequence of such tuples:
\begin{equation}
\begin{aligned}
\chi = & \big((o_1, \phi_{o_1}, a_{\text{low}}^{(o_1, \phi_{o_1})}, 1), (o_2, \phi_{o_2}, a_{\text{low}}^{(o_2, \phi_{o_2})}, 2), \\
 & \dots, (o_n, \phi_{o_n}, a_{\text{low}}^{(o_n, \phi_{o_n})}, n) \big),
\label{eq:sequence}
\end{aligned}
\end{equation}
which serves as the basis for generating a high-level action in the form of a task primitive sequence.

% However, $\chi$ does not account for dependencies among items due to precedence constraints. Some items in $\mathcal{N}$ may be occluded and thus inaccessible until others are packed, yet still appear earlier in the tree. As a result, robots may be unable to execute $\chi$ in order. To resolve this, we reorder $\chi$ into $\tilde{\chi}$, a job queue containing the same items in a different order.
% The number of $\tilde{\chi}$ is the same with the number of leaf vertices in $\mathcal{T}$ so we have $\mathbb{X} = \{\tilde{\chi}_1, \tilde{\chi}_2, \dots\}$. For each $\tilde{\chi}$, we evaluate via forward simulation by computing its score $\mu(\tilde{\chi})$ and utilization ratio $util(\tilde{\chi})$, and generate a corresponding task sequence $a_{\tilde{\chi}}$. The score $\mu(\tilde{\chi})$ is defined as the sum of rewards from executing $a_{\text{low}}^{(o,\phi)}$ in $s_{\text{low}}^{(o,\phi)}$, according to Eq.~\eqref{eq:r_low}. The task sequence $a_{\chi^*}$ generated from the highest-scoring $\chi^*$ is then selected and assigned to $a_{\text{high}}$. The full procedure is detailed in Sec.~\ref{sec:high}.

However, $\chi$ does not account for precedence constraints among items.
Some items in $\mathcal{N}$ are occluded and inaccessible until others are packed, yet they can still appear earlier in the tree.
As a result, robots may be unable to execute $\chi$ in the given order.
To resolve this, we reorder $\chi$ into $\tilde{\chi}$, a job queue that contains the same items in a different order.
Each leaf in the search tree $\mathcal{T}$ produces one such sequence, forming the set $\mathbb{X} = {\tilde{\chi}1, \tilde{\chi}2, \dots}$.
Each $\tilde{\chi}$ is evaluated via forward simulation by computing its score $\mu(\tilde{\chi})$ and bin utilization ratio $\mathit{util}(\tilde{\chi})$, and the corresponding task sequence $a{\tilde{\chi}}$ is generated.
The score $\mu(\tilde{\chi})$ is defined as the cumulative reward obtained by executing $a_{\text{low}}^{(o,\phi)}$ in $s_{\text{low}}^{(o,\phi)}$, as described in \eqref{eq:r_low}.
The task sequence $a_{\chi^*}$ corresponding to the highest-scoring $\chi^*$ is then selected and assigned to $a_{\text{high}}$.
Based on this description about the relationship between the low- and high-level, the full procedure of the high-level is detailed in Sec.~\ref{sec:high}.

\subsection{High-Level: Tree Expansion and Action Generation}
\label{sec:high}

\begin{figure}[!t]

\centerline{\includegraphics[width=\columnwidth]{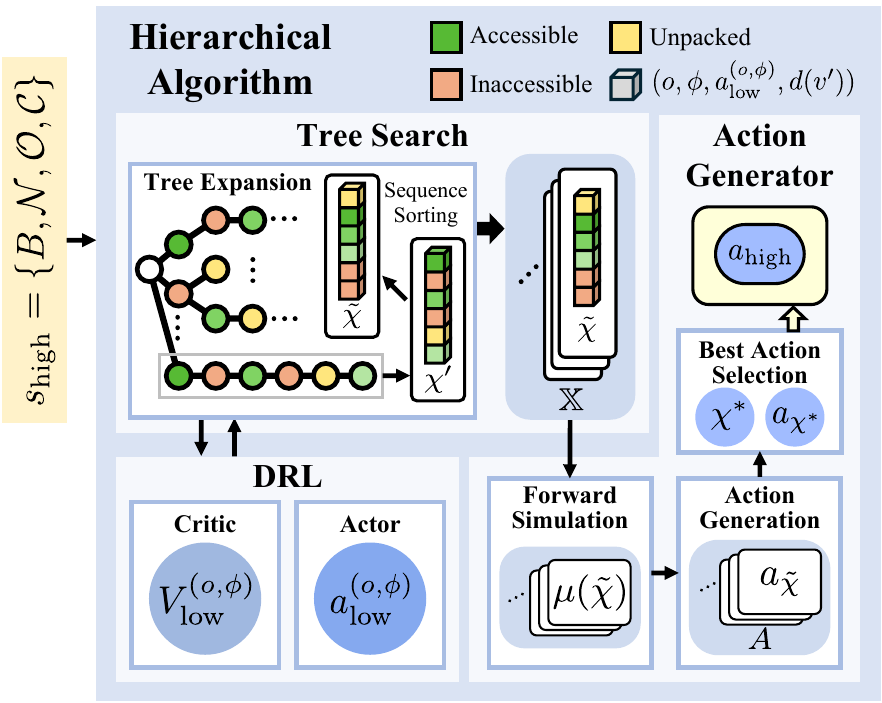}}
    \caption{Hierarchical algorithm. Each candidate sequence $\chi'$ is reordered into $\tilde{\chi}$, evaluated via forward simulation, and the best one is selected to generate the high-level action $a_{\text{high}}$.}
    \label{fig:hierarchical}
    % \vspace{-10pt}
\end{figure}

% We present the high-level search procedure in the hierarchical bin packing algorithm, from tree construction to the selection of the final task sequence. We also describe the repacking process for improved utilization and our strategy for distributing tasks between two manipulators to enable efficient collaborative packing.
We present the high-level procedure of the hierarchical bin packing algorithm, from tree construction to the generation of the final task sequence. The overall process is illustrated in Fig.~\ref{fig:hierarchical}. In addition, we describe the repacking strategy for improved bin utilization.

\subsubsection{High-Level Search Algorithm}
\label{sec:highlevel}

\begin{algorithm}
{
\footnotesize
    \caption{\textsc{High-Level Search}}
    \label{alg:framework2}
    \textbf{Input}: $s_{\text{high}}, \textit{useRepack}, \textit{requireFullPack}$ \\
    \textbf{Output}: $a_{\text{high}}$
    \begin{algorithmic}[1]
        \STATE $v_\text{root} \gets s_{\text{high}}$  \label{line:1-1}\codecomment{Initialization}
        \STATE \mbox{$\mathbb{X}, \textit{solved} \gets$ $\textsc{TreeExpansion}(v_\text{root},\emptyset,(),0,0,\textbf{false})$}  \label{line:1-2}
        \IF{$\mathbb{X} = \emptyset$}  \label{line:1-3}
            \STATE $a_\text{high} \gets (\texttt{terminate})$  \label{line:1-4}
            \STATE $util \gets$ CurrentUtilizationRatio  \label{line:1-4_1}
        \ELSE  \label{line:1-5}
            \STATE $A \gets$ \textsc{SimulationAndGeneration}($\mathbb{X}$)  \label{line:1-6}
            \STATE $\chi^*, a_\text{high} \gets$ \textsc{BestActionSelection}($\mathbb{X}, A$)  \label{line:1-7}
            \STATE $util \gets util(\chi^*)$   \label{line:1-7_1}
        \ENDIF  \label{line:1-8}
        % \STATE $r \gets r_\text{high}(s_\text{high},a_\text{high})$ \label{line:1-16} \\
        \codecommentline{Repacking process}
        \IF{$(\mathbb{X} = \emptyset$ \OR NoPositionAction is in $\chi^*)$ \AND \textit{useRepack}}  \label{line:1-9} %Repack을 시도하고자 한다면
            % \STATE $a^*, RepackSuccess \gets$ \textsc{RepackTrial}($v_\text{root}, r, \textit{requireFullPack}$) \label{line:1-10}
            \STATE $a^*, RepackSuccess \gets$ \textsc{RepackTrial}($v_\text{root}, util, \textit{requireFullPack}$) \label{line:1-10}
            \IF{$RepackSuccess$}  \label{line:1-11}
                \STATE $a_\text{high} \gets a^*$  \label{line:1-12}
            \ENDIF  \label{line:1-13}
        \ENDIF  \label{line:1-14}
        \RETURN $a_\text{high}$  \label{line:1-15}
    \end{algorithmic}
}
\end{algorithm}

% The input to Alg.~\ref{alg:framework2} includes the initial high-level state $s_{\text{high}}$, a boolean flag \textit{useRepack} indicating whether to attempt repacking, and a flag \textit{requireFullPack} to distinguish between the following two modes in Alg.~\ref{alg:repacking}:
% (i) If the goal is to find an optimal solution and sub-optimal ones are not of interest, the full pack mode is enabled. For instances where 100\% bin utilization is achievable, Alg.~\ref{alg:repacking} continues the search until such a solution is found.
% % In practice, a 3-second time budget is often sufficient at each additional search step to find a solution.
% (ii) If the goal is to find the best solution within a given time limit, the full pack mode is disabled.
% Alg.~\ref{alg:repacking} then performs progressive refinement and returns the best solution found.
% % Even with a short budget (e.g., 1--2 seconds) for additional search, this mode significantly improves packing quality compared to not using repacking.
% The output of Alg.~\ref{alg:framework2} is the high-level action $a_\text{high}$, a sequence of task primitives for bin packing.

The input to Alg.~\ref{alg:framework2} includes the initial high-level state $s_{\text{high}}$, a boolean flag \textit{useRepack} indicating whether to attempt repacking, and a flag \textit{requireFullPack} to distinguish between the following two modes:  
(i) \textit{Full-pack} mode: This mode focuses solely on finding an optimal solution in which all known items are packed, and sub-optimal ones are not of interest. For instances where 100\% bin utilization is achievable, Alg.~\ref{alg:repacking} continues the search until such a solution is found.  
(ii) \textit{Any-time} mode: In this mode, the algorithm performs progressive refinement and returns the best solution found within a limited time budget.  
The output of Alg.~\ref{alg:framework2} is the high-level action $a_\text{high}$, a sequence of task primitives for bin packing.

Alg.~\ref{alg:framework2} initializes the root vertex $v_\text{root}$ with $s_{\text{high}}$ (line~\ref{line:1-1}).
A tree search is then performed via the recursive function \textsc{TreeExpansion} (detailed in Alg.~\ref{alg:pack}), which expands the search tree and returns the set of candidate packing sequences $\mathbb{X}$ along with a boolean flag $\textit{solved}$ indicating whether a solution achieving 100\% bin utilization has been found (line~\ref{line:1-2}). The \textit{solved} remains false unless 100\% utilization is reached, even if all known items can be packed.

% If no valid solution is found ($\mathbb{X} = \emptyset$), Alg.~\ref{alg:framework2} sets $a_\text{high}$ to $\texttt{terminate}$ to end the episode and $util$, the max utilization ratio found until now, is assigned the current utilization ratio (lines~\ref{line:1-3}--\ref{line:1-4_1}).
% Otherwise, the sequences $\tilde{\chi} \in \mathbb{X}$ are evaluated, and the candidate high-level action set $A$ is generated via \textsc{SimulationAndGeneration} (line~\ref{line:1-6}). Then, \textsc{BestActionSelection} selects the optimal sequence $\chi^*$ and assigns its corresponding action to $a_\text{high}$ (line~\ref{line:1-7}). Both procedures are described in Sec.~\ref{sec:action}.
% After obtaining $a_\text{high}$, $util$ is assigned the utilization ratio $util(\chi^*)$ computed from $\chi^*$ (line~\ref{line:1-7_1}).

If no valid solution is found ($\mathbb{X} = \emptyset$), Alg.~\ref{alg:framework2} sets $a_\text{high}$ to $\texttt{terminate}$ to end the episode, and assigns $util$, the highest achievable utilization ratio identified so far, as the current value (lines~\ref{line:1-3}--\ref{line:1-4_1}).
Otherwise, each sequence $\tilde{\chi} \in \mathbb{X}$ is evaluated, and the candidate high-level action set $A$ is generated through \textsc{SimulationAndGeneration} (line~\ref{line:1-6}).
Then, \textsc{BestActionSelection} selects the best sequence $\chi^*$ and sets its action as $a_\text{high}$ (line~\ref{line:1-7}).
Both procedures are detailed in Sec.~\ref{sec:action}.
After $\chi^*$ is determined, $util$ is assigned as $\mathit{util}(\chi^*)$, the utilization ratio computed from $\chi^*$ (line~\ref{line:1-7_1}).

If the initial search yields no valid placements (i.e., $\mathbb{X} = \emptyset$) or the selected sequence includes no-position actions, with repacking enabled, Alg.~\ref{alg:framework2} invokes \textsc{RepackTrial} (Alg.~\ref{alg:repacking}) to iteratively unpack and re-search within the additional time (lines~\ref{line:1-9}--\ref{line:1-10}). 
If repacking succeeds, the resulting action $a^*$ replaces $a_\text{high}$ (lines~\ref{line:1-11}--\ref{line:1-12}).
Finally, $a_\text{high}$ is returned (line~\ref{line:1-15}).

\subsubsection{Tree Construction}
\label{sec:tree_construction}

\begin{algorithm}
{
\footnotesize
    \caption{\textsc{TreeExpansion}}
    \label{alg:pack}
    \textbf{Input}: $v, \mathbb{X}, \chi, d, n, \textit{requireFullPack}$\\
    \textbf{Output}: $\mathbb{X}, \textit{solved}$
    \begin{algorithmic}[1]
        \STATE $\mathcal{O} \gets ();$ $n=0?\ stop = \text{\bf{true}}:stop = \text{\bf{false}}$ \codecomment{Initialization}  \label{line:2-1} 
        \FOR{each $o \in \mathcal{N} \cup \mathcal{C}$}  \label{line:2-2}
            \FOR{each $\phi \in \{0,\pi/2\}$}  \label{line:2-3}
                \STATE $a_{\text{low}}^{(o,\phi)} \gets \pi_{\text{low}}(s_\text{low}^{(o,\phi)})$ and compute $r(s^{(o,\phi)}_\text{low},a_{\text{low}}^{(o,\phi)})$   \label{line:2-4}
                % \IF{$o$ is accessible \AND $a_{\text{low}}^{(o,\phi)} \neq W \cdot H$}  \label{line:2-5}
                \IF{$o$ is accessible \AND $a_{\text{low}}^{(o,\phi)}$ is not NoPositionAction}  \label{line:2-5}
                    \STATE $\textit{stop} \gets \FALSE$  \label{line:2-6}
                \ENDIF  \label{line:2-7}
                \STATE $\textsc{Append}(\mathcal{O},(o, \phi))$  \label{line:2-8}
            \ENDFOR  \label{line:2-9}
            % \IF{Repacking is included \AND $a_{\text{low}}^{(o,0)} = W \cdot H$ \AND $a_{\text{low}}^{(o,\pi/2)} = W \cdot H$}  \label{line:2-10}
            \IF{\textit{requireFullPack} \AND $(a_{\text{low}}^{(o,0)}, a_{\text{low}}^{(o,\pi/2)} \text{ are NoPositionAction})$}  \label{line:2-10}
                \RETURN $\mathbb{X}, \FALSE$  \label{line:2-11}
            \ENDIF  \label{line:2-12}
        \ENDFOR  \label{line:2-13}
        \IF{$stop$}  \label{line:2-14}
            \RETURN $\mathbb{X}, \FALSE$  \label{line:2-15}
        \ENDIF  \label{line:2-16}
        \STATE $\mathcal{O} \gets$ \textsc{RewardSorting}($\mathcal{O}$)  \label{line:2-17}
        \STATE $\mathcal{O} \gets$ \textsc{Selection}($\mathcal{O}$)  \label{line:2-18}
        \FOR{each $(o,\phi) \in \mathcal{O}$}  \label{line:2-19}
            \STATE $\chi^{\prime} \gets \textsc{Append}(\chi,\ (o, \phi, a_{\text{low}}^{(o,\phi)}, d))$  \label{line:2-20}
            % \IF{$a_{\text{low}}^{(o,\phi)} \neq W \cdot H$}  \label{line:2-21}
            \IF{$a_{\text{low}}^{(o,\phi)}$ is not NoPositionAction}  \label{line:2-21}
                \STATE $n^{\prime} \gets$ \textsc{NodeCountUpdate}($n, o$)  \label{line:2-22}
                \STATE $v' \gets$ \textsc{ChildNodesGeneration}($v, o, \phi, a_{\text{low}}^{(o,\phi)}$)  \label{line:2-23}
                \STATE $\textit{full} \gets$ \textsc{BinFullCheck}($B^\prime$)  \label{line:2-24}
            \ENDIF  \label{line:2-25}
            \IF{$a_{\text{low}}^{(o,\phi)}$ is NoPositionAction \OR $\mathcal{N}^{\prime} \cup \mathcal{C}^{\prime} = \emptyset$ \OR \textit{full}}  \label{line:2-26}
                \STATE $\tilde{\chi} \gets$ \textsc{SequenceSorting}($\chi^{\prime}$)  \label{line:2-27}
                \STATE $\mathbb{X} \gets \mathbb{X} \cup \{\tilde{\chi}\}$  \label{line:2-28}
                \IF{\textit{requireFullPack} \AND \textit{full}} \label{line:2-29}
                    \RETURN $\mathbb{X}, \TRUE$  \label{line:2-30}
                \ENDIF  \label{line:2-31}
                \STATE \textbf{continue}  \label{line:2-32}
            \ENDIF  \label{line:2-33}
            \STATE $\mathbb{X}, \textit{solved} \gets$ \\
                    \hfill \textsc{TreeExpansion}($v',\mathbb{X},\chi^{\prime},d+1,n^{\prime},requireFullPack$) \label{line:2-34}
            \IF{$\textit{solved}$}  \label{line:2-35}
                \RETURN $\mathbb{X}, \TRUE$  \label{line:2-36}
            \ENDIF  \label{line:2-37}
        \ENDFOR  \label{line:2-38}
        \RETURN $\mathbb{X}, \FALSE$  \label{line:2-39}
    \end{algorithmic}
}
\end{algorithm}

Alg.~\ref{alg:pack} recursively expands the search tree to find an optimal packing sequence for the hierarchical bin packing problem. Alg.~\ref{alg:pack} follows a DFS-BS strategy, combining depth-first expansion with selective branching. Promising branches are prioritized based on heuristic scores, and the search width is adjusted to balance exploration and computational efficiency.
Alg.~\ref{alg:pack} takes as input the current state vertex $v$, the set of tuple sequences $\mathbb{X}$, the current sequence $\chi$ (which contains all tuples from the root to $v$), the current depth $d$, the number $n$ of accessible items in previous nodes, and the boolean flag \textit{requireFullPack}.

Alg.~\ref{alg:pack} begins by initializing $\mathcal{O}$, which stores candidate item–orientation pairs $(o, \phi)$.
The stopping flag $\textit{stop}$ is activated if no accessible items have been considered (line~\ref{line:2-1}).

Each unprocessed item $o$ is considered in both orientations $\phi \in \{0, \pi/2\}$ (lines~\ref{line:2-2}--\ref{line:2-3}).
For each $\phi$, Alg.~\ref{alg:pack} selects the low-level action $a_{\text{low}}^{(o,\phi)}$ using $\pi_{\text{low}}$ and evaluates its reward $r(s^{(o,\phi)}_\text{low}, a_{\text{low}}^{(o,\phi)})$ (line~\ref{line:2-4}).
If an accessible item has at least one valid placement, $\textit{stop}$ is deactivated (lines~\ref{line:2-5}--\ref{line:2-6}).
Considered  $(o, \phi)$ pair is added to $\mathcal{O}$ (line~\ref{line:2-8}).
If the full-pack mode is enabled and an item $o$ yields no-position actions in both orientations, the current branch is immediately aborted (lines~\ref{line:2-10}--\ref{line:2-11}).
This avoids exploring paths that cannot lead to full utilization. If no valid placements are found for any accessible item, the current branch is aborted (lines~\ref{line:2-14}--\ref{line:2-15}).

After evaluating all candidates, $\mathcal{O}$ is sorted by \textsc{RewardSorting} (line~\ref{line:2-17}).
This function prioritizes candidates by descending reward $r(s^{(o,\phi)}_\text{low}, a_{\text{low}}^{(o,\phi)})$, and in case of a tie, by descending item size.
Placements with higher rewards improve bin utilization, and considering larger items earlier improves overall space usage, as larger items are more difficult to fit in later stages.
Then, \textsc{Selection} selects top-ranked candidates based on reward values (line~\ref{line:2-18}).
By controlling the branching factor, this strategy enhances both exploration quality and computational efficiency.
Unselected candidates remain in $\mathcal{N} \cup \mathcal{C}$ and are reconsidered at deeper levels.
If all candidates yield no-position actions, one is retained to preserve the current partial sequence.

Alg.~\ref{alg:pack} iterates over the sorted-selected $\mathcal{O}$ to generate child nodes (line~\ref{line:2-19}).
Each element of $\chi$ is a tuple of an item, its orientation, the positional action, and the node depth, as defined in \eqref{eq:sequence}.
These values are obtained from the current recursion context and each candidate in $\mathcal{O}$.
% A new tuple is appended to $\chi$ for each candidate, yielding $|\mathcal{O}|$ extended sequences (line~\ref{line:2-20}).
A new tuple is appended to $\chi$ for each candidate (line~\ref{line:2-20}).
If $a_{\text{low}}^{(o,\phi)}$ is valid, a child node $v'$ is generated, and the count $n'$ of accessible packable items is updated.
The updated bin $B'$ is checked for full occupancy, and the result is stored in the \textit{full} flag (lines~\ref{line:2-22}--\ref{line:2-24}).
If the item cannot be placed, all items have been processed, or the bin is full, Alg.~\ref{alg:pack} sorts the current sequence $\chi'$ into $\tilde{\chi}$ and adds it to the solution set $\mathbb{X}$ (lines~\ref{line:2-26}--\ref{line:2-28}).
The sorting prioritizes unpacked items, followed by accessible and then inaccessible ones.
Among inaccessible items, arrival order determines their ordering in $\tilde{\chi}$.
If both \textit{requireFullPack} and \textit{full} are true, Alg.~\ref{alg:pack} terminates early by returning true (lines~\ref{line:2-29}--\ref{line:2-30}).
Otherwise, since the condition in line~\ref{line:2-26} indicates no further expansion is needed, Alg.~\ref{alg:pack} skips the recursive call and proceeds to the next branch (line~\ref{line:2-32}).
Unless the condition in line~\ref{line:2-26} is satisfied, Alg.~\ref{alg:pack} recursively calls \textsc{TreeExpansion} to explore the next depth (line~\ref{line:2-34}).
If the recursive call returns with \textit{solved} set to true, Alg.~\ref{alg:pack} terminates (lines~\ref{line:2-35}--\ref{line:2-36}).
After exploring all candidates Alg.~\ref{alg:pack} returns the set $\mathbb{X}$ (line~\ref{line:2-39}).

\subsubsection{Action Generation and Forward Simulation}
\label{sec:action}

\begin{figure}
\centering
    \includegraphics[width=1.0\linewidth]{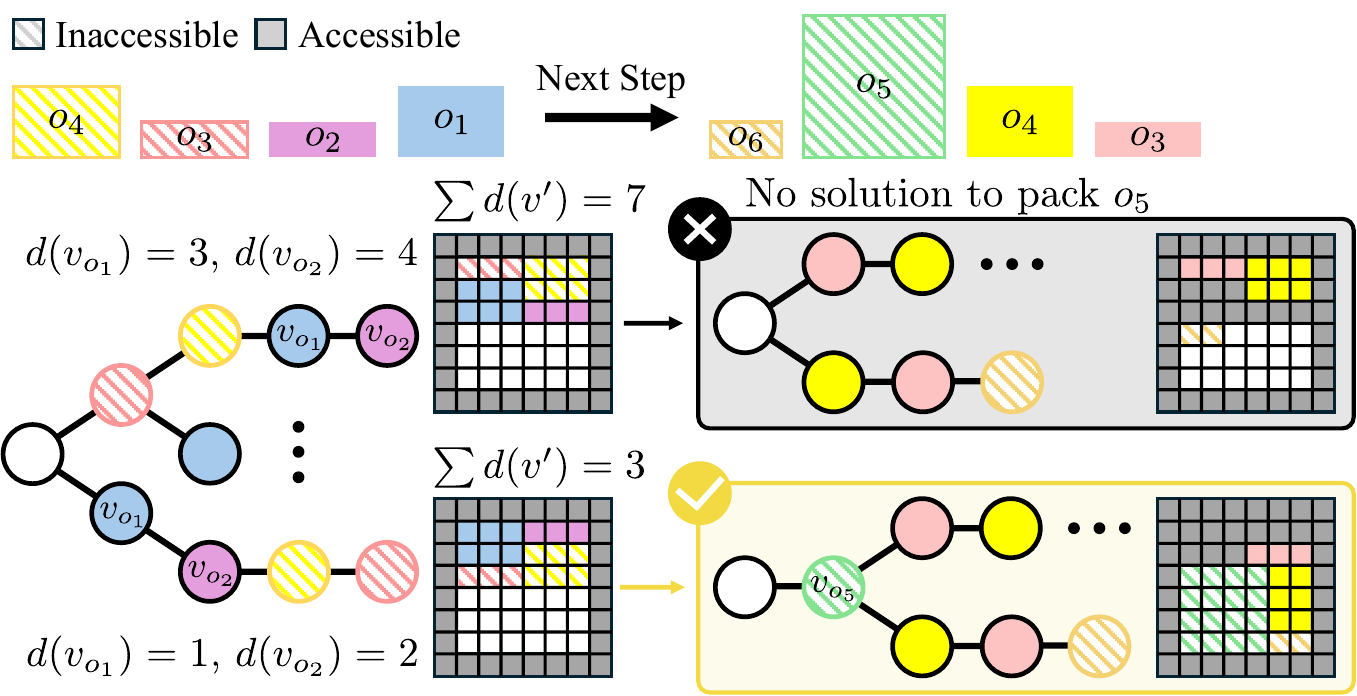}
    \caption{
    Solution prioritization by total tree depth of accessible items.
    Among solutions yielding the same bin configuration, those that place accessible items (e.g., $o_1$, $o_2$) earlier tend to produce more compact arrangements, improving adaptability to future arrivals.
    }
    % \vspace{-10pt}
\label{fig:depth}
\end{figure}

Once the tree is fully expanded, actions are generated and forward simulation is performed for each $\tilde{\chi} \in \mathbb{X}$, constructed by Alg.~\ref{alg:pack}.
The forward simulation is designed to evaluate multiple candidate solutions generated by the tree search, determining which solution $\tilde{\chi}$ leads to better future outcomes while ensuring the feasibility of the packing sequence and placement positions.
Through forward simulation, the score $\mu(\tilde{\chi})$ is calculated from an initial value of zero.
Each element of $\tilde{\chi}$ is a tuple $(o, \phi, a^{(o,\phi)}_\text{low}, d(v'))$.
Tuples with valid positional actions are sequentially converted into \texttt{pack}$(o, \phi, x_o, y_o)$ commands.
If $o$ is an unpacked or accessible item, the \texttt{pack} action is executed in the simulation, and the corresponding low-level reward from $a^{(o,\phi)}_\text{low}$ is added to $\mu(\tilde{\chi})$.
In conveyor-based environments, if an item is removed from $\mathcal{N}$ via \texttt{pack}, subsequent items advance, potentially making previously inaccessible items accessible.
If the action is a no-position action or the item remains inaccessible, the item is skipped in the simulation.
The simulation terminates when either all accessible items result in no-position actions or all tuples in $\tilde{\chi}$ have been processed.
Those containing more skipped or infeasible items than others are naturally filtered out during the evaluation process, since they yield lower cumulative scores.
Then, $util(\tilde{\chi})$, the final utilization ratio of $\tilde{\chi}$, is computed.

As the forward simulation progresses, \texttt{pack} actions are appended to the corresponding high-level action $a_{\tilde{\chi}}$.
All \texttt{pack} actions for unpacked items are included.
For accessible items not yet packed, at most as many \texttt{pack} primitives are added as the number of available robots.
Other accessible items with valid positional actions are executed in the simulation but excluded from $a_{\tilde{\chi}}$ at the current step.
Among accessible items, inclusion in $a_{\tilde{\chi}}$ is based on the depth value $d(v')$.
In the proposed synchronized dual-arm setting, two objects are selected and assigned to the respective robots.
The item with the smallest $d(v')$ accessible by at least one robot is selected first.
If only one robot can access it, the other is assigned the item with the next smallest $d(v')$ it can access.
If both robots can access the same item, the second smallest $d(v')$ item accessible by either is selected next.
If the next conveyor step yields no packable items, a \texttt{terminate} primitive is appended to $a_{\tilde{\chi}}$.

After the forward simulation, the sequence with the highest score $\mu(\tilde{\chi})$ is selected.
If multiple sequences yield the same score, the one with the highest utilization ratio $util(\tilde{\chi})$ is chosen.
If a tie still remains, the sequence with the smallest total $d(v')$ for items included in the high-level action is selected,
as prioritizing items with smaller depth values improves adaptability to unknown future arrivals, as illustrated in Fig.~\ref{fig:depth}.

\subsubsection{Repacking Solution Search}

Although Alg.~\ref{alg:pack} finds a high-quality packing solution, it may not reach global optimality in the online BPP setting, where only partial item information is available at decision time.
To address this, we propose an improvement procedure that enhances bin utilization by selectively unpacking and repacking items in different positions or orientations.
Since Alg.~\ref{alg:repacking} starts from a valid solution and can terminate at any time while progressively improving the result, it functions as an anytime algorithm.

Alg.~\ref{alg:repacking} initializes the candidate best action $a^*$ to return, a tuple $a$ to store \texttt{unpack} primitives, and the repacking success flag \textit{RepackSuccess} as false (line~\ref{line:3-1}).
Alg.~\ref{alg:repacking} explores combinations of items to unpack from $\mathcal{I}$, gradually increasing the number of unpacked items $i$ while staying within the time limit (lines\ref{line:3-2}--\ref{line:3-3}).
For each $i$, candidate subsets $\mathcal{U}$ with $i$ elements of packed items $\mathcal{I}$ are considered, prioritizing those that include more recently packed items first, as earlier-packed ones typically form the core of the compact structure.
For each $\mathcal{U}$, the current environment state $(B, \mathcal{N}, \mathcal{I}, \mathcal{C})$ is cloned into $(\hat{B}, \hat{\mathcal{N}}, \hat{\mathcal{I}}, \hat{\mathcal{C}})$, and items $o \in \mathcal{U}$ are sequentially unpacked (lines~\ref{line:3-5}--\ref{line:3-8}).
A new tree search is then performed on the updated state, using the \textit{requireFullPack} flag introduced in Sec.~\ref{sec:highlevel}  (line~\ref{line:3-10}).
If a valid sequence set $\mathbb{X}$ is obtained (line~\ref{line:3-11}), the corresponding high-level actions $A$ are generated, and the best one $(\chi^*, a_{\chi^*})$ is selected (lines~\ref{line:3-12}--\ref{line:3-13}).
The final action is constructed by prepending the unpack operations $a$ to $a_{\chi^*}$ through \textsc{ActionUpdate} (line~\ref{line:3-14}).
If full-pack mode is enabled and $\chi^*$ contains no no-position actions, repacking is considered successful: \textit{RepackSuccess} is set to true, and Alg.~\ref{alg:repacking} terminates (lines~\ref{line:3-15}--\ref{line:3-18}).
If any-time mode is disabled (i.e., \textit{requireFullPack} is false) and the utilization $\mathit{util}(\chi^*)$ exceeds the current best $util$, the solution is accepted as an improvement: $a^*$, \textit{RepackSuccess}, and $util$ are updated (lines~\ref{line:3-19}--\ref{line:3-22}).
Before proceeding to the next unpacking subset, the unpack action list $a$ is reset to prevent interference with future trials (line~\ref{line:3-25}).
The procedure continues examining new combinations until the time limit is reached.
Finally, the best action $a^*$ found during the process, along with the flag \textit{RepackSuccess}, is returned (line~\ref{line:3-29}).

\begin{algorithm}
{
\footnotesize
    \caption{\textsc{RepackTrial}}
    \label{alg:repacking}
    % \textbf{Input}: $v_\text{root}, r, \textit{requireFullPack}$ \\
    \textbf{Input}: $v_\text{root}, util, \textit{requireFullPack}$ \\
    \textbf{Output}: $a^*, RepackSuccess$
    \begin{algorithmic}[1]
        \STATE $a^* \gets (), a \gets (), RepackSuccess$ $\gets \FALSE$  \label{line:3-1}  \codecomment{Initialization}
        \WHILE{time limit not reached}  \label{line:3-2}
            \FOR{$i \gets 1$ to $|\mathcal{I}|$}  \label{line:3-3}
                \FOR{each subset $\mathcal{U}$ of $\mathcal{I}$ with $i$ elements (last-packed-first)} \label{line:3-4}
                    \STATE Clone $B, \mathcal{N}, \mathcal{I}, \mathcal{C}$ to $\hat{B}, \hat{\mathcal{N}}, \hat{\mathcal{I}}, \hat{\mathcal{C}}$  \label{line:3-5}
                    \FOR{each item $o$ in $\mathcal{U}$}  \label{line:3-6}
                        \STATE Execute $\texttt{unpack}(o)$ and update $\hat{v}_\text{root}$ with $\hat{B}, \hat{\mathcal{N}}, \hat{\mathcal{I}}, \hat{\mathcal{C}}$  \label{line:3-7}
                        \STATE $a \gets \textsc{Append}(a,\texttt{unpack}(o))$  \label{line:3-8}
                    \ENDFOR  \label{line:3-9}
                    \STATE $\mathbb{X}, \textit{solved} \gets$ \\
                        \hfill \mbox{$\textsc{TreeExpansion}(\hat{v}_\text{root},\emptyset,(),0,0,\textit{requireFullPack})$}  \label{line:3-10}
                    \IF{$\mathbb{X} \neq \emptyset$}  \label{line:3-11}
                        \STATE $A \gets$ \textsc{SimulationAndGeneration}($\mathbb{X}$)  \label{line:3-12}
                        \STATE $\chi^*, a_{\chi^*} \gets$ \textsc{BestActionSelection}($\mathbb{X}, A$)  \label{line:3-13}
                        \STATE $a_{\chi^*} \gets \textsc{ActionUpdate}(a, a_{\chi^*})$ \label{line:3-14}
                        \IF{\textit{requireFullPack} \AND NoPositionAction is not in $\chi^*$}  \label{line:3-15}
                            \STATE $RepackSuccess$ $\gets \TRUE$  \label{line:3-16}
                            \RETURN $a_{\chi^*}, RepackSuccess$  \label{line:3-18}
                        % \ELSIF{\textbf{not} \textit{requireFullPack} \AND $r < r_\text{high}(s_\text{high},a_{\chi^*})$}  \label{line:3-19}
                        \ELSIF{\textbf{not} \textit{requireFullPack} \AND $util < util(\chi^*)$}  \label{line:3-19}
                            \STATE $RepackSuccess$ $\gets \TRUE$  \label{line:3-20}
                            % \STATE $a^* \gets a_{\chi^*}, r \gets r_\text{high}(s_\text{high},a_{\chi^*})$  \label{line:3-22}
                            \STATE $a^* \gets a_{\chi^*}, util \gets  util(\chi^*)$  \label{line:3-22}
                            % \IF{\textit{solved}}  \label{line:3-22}
                            %     \RETURN $a_\text{high}, RepackSuccess$  \label{line:3-23}
                            % \ENDIF  \label{line:3-24}
                        \ENDIF  \label{line:3-23}
                    \ENDIF  \label{line:3-24}
                    \STATE Reset $a \gets ()$  \label{line:3-25}
                \ENDFOR  \label{line:3-26}
            \ENDFOR  \label{line:3-27}
        \ENDWHILE  \label{line:3-28}
        \RETURN $a^*, RepackSuccess$  \label{line:3-29}
    \end{algorithmic}
}
\end{algorithm}

\subsection{Task Allocation and Atomic Action Sequencing}

Until now, action primitives have been generated without being assigned to specific robots. While single-arm systems have no ambiguity, dual-arm systems require explicit task distribution.
\texttt{pack} primitives for newly arrived items are allocated to the robot closest in Euclidean distance.
The primitives \texttt{unpack} are assigned alternately, starting with the less-loaded robot.
\texttt{pack} for unpacked items is assigned to the same robot that executed the corresponding \texttt{unpack}, with repacking done in reverse order of unpacking.
Robots resume packing new items after completing all unpacking and repacking tasks.

To generate feasible motions for executing the assigned task primitives, we decompose them into atomic actions.
As dual-arm motion planning in a shared task space is an inherently complex problem, we sequence actions in a synchronized manner rather than planning full asynchronous motions.
This strategy reduces planning time at the expense of execution time.
Our \textit{atomic action sequencing} guarantees that robots do not operate in overlapping task space regions simultaneously.

In our environment, we define three task subspaces: the bin, the conveyor belt, and the buffer space of each robot. Both \texttt{pack} and \texttt{unpack} are decomposed into two atomic actions: \texttt{pick} and \texttt{place to (location)}.
\begin{itemize}
    \item \texttt{pack}: \texttt{pick} from the conveyor or buffer, then \texttt{place to bin}.
    \item \texttt{unpack}: \texttt{pick} from the bin, then \texttt{place to buffer}.
\end{itemize}
We also define two additional atomic actions: \texttt{standby}, where a robot waits, and \texttt{ready}, where it returns to its initial pose after completing all assigned tasks.

Atomic action sequencing is constructed as follows:
\begin{enumerate}[label=(\roman*)]
% \small
    \item Assign \texttt{standby} to the robot with fewer tasks.
    \item Add \texttt{pick} and \texttt{place to buffer} for all \texttt{unpack} from the first without a corresponding \texttt{pack} to the last with one.
    % \item If the last \texttt{unpack} has corresponding \texttt{pack}, omit \texttt{place to buffer}.
    % \item In the corresponding \texttt{pack} to the last \texttt{unpack}, omit \texttt{pick} and include only \texttt{place to bin}.
    \item If the last \texttt{unpack} is followed by a corresponding \texttt{pack}, omit \texttt{place to buffer} and \texttt{pick}, yielding \texttt{pick} – \texttt{place to bin}.
    \item For all remaining \texttt{pack}, add \texttt{pick} and \texttt{place to bin}.
    \item Conclude with a \texttt{ready} action for each robot.
\end{enumerate}

Adding \texttt{standby} to one robot ensures that the two tobots do not operate in overlapping regions simultaneously, simplifying dual-arm motion planning.
Any off-the-shelf motion planner can be used; we adopt an efficient variant suitable for multi-arm coordination.
% To execute motion planning for the dual-arm system, we employ StopNGo~\cite{stopngo}.

\begin{table}[t]
% \footnotesize
\centering
\caption{Scenario configurations with varying robot and item setups.}
\label{table:scenario}

\scalebox{0.95}
{
\begin{tabular}{|c|ccc|ccc|}
\hline
\textbf{Scenario}                                                                           & \multicolumn{1}{c|}{\begin{tabular}[c]{@{}c@{}}S-R1\\ A1\end{tabular}} & \multicolumn{1}{c|}{\begin{tabular}[c]{@{}c@{}}S-R5\\ A1\end{tabular}} & \begin{tabular}[c]{@{}c@{}}S-R5\\ A3\end{tabular} & \multicolumn{1}{c|}{\begin{tabular}[c]{@{}c@{}}D-R2\\ A2O2\end{tabular}} & \multicolumn{1}{c|}{\begin{tabular}[c]{@{}c@{}}D-R5\\ A2O2\end{tabular}} & \begin{tabular}[c]{@{}c@{}}D-R5\\ A3O1\end{tabular} \\ \hline
\textbf{Fig.~\ref{fig:scenarios}}                                                              & \multicolumn{1}{c|}{(a)}                                               & \multicolumn{1}{c|}{(b)}                                               & (c)                                               & \multicolumn{1}{c|}{(d)}                                                 & \multicolumn{1}{c|}{(e)}                                                 & (f)                                                 \\ \hline
\textbf{Num. of Robots}                                                                     & \multicolumn{3}{c|}{1}                                                                                                                                                                              & \multicolumn{3}{c|}{2}                                                                                                                                                                                    \\ \hline
\textbf{\begin{tabular}[c]{@{}c@{}}Max Num. of\\ Recognized Items\end{tabular}}             & \multicolumn{1}{c|}{1}                                                 & \multicolumn{1}{c|}{5}                                                 & 5                                                 & \multicolumn{1}{c|}{2}                                                   & \multicolumn{1}{c|}{5}                                                   & 5                                                   \\ \hline
\textbf{\begin{tabular}[c]{@{}c@{}}Num. of Accessible\\ Items per Robot\end{tabular}}       & \multicolumn{1}{c|}{1}                                                 & \multicolumn{1}{c|}{1}                                                 & 3                                                 & \multicolumn{1}{c|}{2}                                                   & \multicolumn{1}{c|}{2}                                                   & 3                                                   \\ \hline
\textbf{\begin{tabular}[c]{@{}c@{}}Num. of Items\\ in Overlapping\\ Workspace\end{tabular}} & \multicolumn{1}{c|}{-}                                                 & \multicolumn{1}{c|}{-}                                                 & -                                                 & \multicolumn{1}{c|}{2}                                                   & \multicolumn{1}{c|}{2}                                                   & 1                                                   \\ \hline
\end{tabular}
}
% \vspace{-10pt}
\end{table}

\section{Experiments}

We evaluate our hierarchical bin packing framework using a hybrid architecture, integrating task planning, DRL, and motion planning modules across two systems.
The algorithm runs on a machine with an AMD Ryzen 7 9700X CPU, 64~GB RAM, and a GeForce RTX 4080 SUPER GPU.
Simulation and motion planning are executed on a separate system with an AMD Ryzen 7 5800X CPU, 32~GB RAM, and a GeForce RTX 3070 Ti GPU.
Modules communicate via ROS1~\cite{roswebsite} and TCP/IP to synchronize the interaction between the planner and simulation.
The DRL agent is implemented in PyTorch~\cite{paszke2019pytorch}, and the simulation environment is developed in Unity~\cite{unity_robotics_hub}, providing physical dynamics.
All training and evaluation experiments are conducted in a $10 \times 10$ bin environment.
Considering the placement margin, robot positioning accuracy, and camera perception error, the 10 × 10 grid resolution is sufficiently precise for stable packing and is therefore adopted as the representative configuration for experimental validation. % review 1-4

\subsection{Dataset}

% We construct two types of datasets: \textit{Random} instances, used only for testing, and \textit{100\% set} instances, used for both training and testing.

% In the \textit{Random} instance, items are generated by sampling the width and height of each item uniformly, with each dimension limited to at most half the bin size.
% A large number of sets are included to ensure sufficient diversity and test coverage.

% The \textit{100\% set} instances are constructed to train the RL agent and evaluate the algorithm.
% It consists of item sets where full bin utilization (100\%) is achievable.
% These sets enable the agent to learn optimal packing strategies and serve as a benchmark to test whether repacking can fully utilize the bin space.
% To construct each set, item dimensions are randomly sampled and placed at feasible locations to ensure compact, valid placements.
% Oversized items are avoided by considering remaining space during sampling.
% Once the bin is fully packed, the item order is shuffled to simulate online arrival.
% In total, 9.2 million item sets are generated across \textit{100\% set} instances: 7.2 million (78\%) are used to train the A3C worker networks, and 2 million (22\%) are reserved exclusively for evaluating the global network.

We construct two types of datasets: \textit{Random} instances, used only for testing, and \textit{100\% set} instances, used for both training and testing.

In the \textit{Random} instance, items are generated by independently sampling each dimension uniformly: $w_o \sim \mathcal{U}(1, \lfloor W/2 \rfloor)$ and $h_o \sim \mathcal{U}(1, \lfloor H/2 \rfloor)$, ensuring no single item exceeds half the bin size in either dimension.

% 100\% Set Instances consist of item sets where full bin utilization (100\%) is achievable, enabling the agent to learn optimal packing strategies and serving as a benchmark to test whether repacking can fully utilize the bin space. The construction procedure is detailed in Alg.~\ref{alg:dataset_generation}. Briefly, item dimensions are sampled from Gaussian-weighted probabilities with standard deviation $\sigma=2$, which reduces the probability of generating items with width or height of 1 in early stages. This prevents an overabundance of such small items that would inevitably arise later when filling remaining gaps. Items are greedily placed at positions that maximize a bottom-left heuristic score combined with edge contact rewards. Oversized items are rejected by considering remaining space during sampling. Once the bin is fully packed, the item order is shuffled to simulate online arrival.

\textit{100\% Set} instances consist of item sets where full bin utilization (100\%) is achievable, enabling the agent to learn optimal packing strategies and serving as a benchmark to test whether repacking can fully utilize the bin space. The construction procedure is detailed in Alg.~\ref{alg:dataset_generation}. Item dimensions are sampled from Gaussian-weighted probabilities (standard deviation $\sigma=2$) over width and height, which reduces the probability of generating items with dimension 1 in early stages.
This design prevents an overabundance of such small or elongated items that would inevitably arise later when filling remaining gaps. For each sampled item, we compute edge-contact rewards for all feasible placement positions and greedily select the position that maximizes this reward, with random tie-breaking when multiple positions share the maximum value. Oversized items that cannot fit in the remaining space are rejected and resampled. Once the bin is full or only one cell remains empty, the process is terminated. In the case where one cell remains empty, a $1 \times 1$ item is added to achieve exact 100\% fill. Finally, the item order is shuffled to simulate online arrival.
In total, 9.2 million item sets are generated: 7.2 million (78\%) are used to train the A3C worker networks, and 2 million (22\%) are reserved exclusively for evaluating the global network.

\begin{algorithm}[t]
\footnotesize
\caption{100\% Set Instance Generation}
\label{alg:dataset_generation}
\begin{algorithmic}[1]
\REQUIRE Bin size $W \times H$, Gaussian std $\sigma=2$
\ENSURE Item set $\mathcal{S}$ with 100\% bin utilization
\STATE $\mathcal{S} \gets \emptyset$, $\text{area} \gets 0$
\STATE Initialize empty bin $B$ of size $W \times H$
\STATE $\max_w \gets 0$, $\max_h \gets 0$
\STATE $p_{w} \gets \text{GaussianProb}(W, \sigma)$
\STATE $p_{h} \gets \text{GaussianProb}(H, \sigma)$
\WHILE{$\text{area} < W \times H - 1$}
    \REPEAT
        \REPEAT
            \STATE Sample $w_o \sim p_{w}$, $h_o \sim p_{h}$
        \UNTIL{$(w_o + h_o \leq W + H - \max_w - \max_h)$ AND \\
            \hspace{17pt} $(\text{area} + w_o \cdot h_o \leq W \times H)$}
        \STATE Initialize $R[y,x] \gets -\infty$ for all $(x,y)$
        \STATE Compute $R[y,x] \gets$ \\
            \hfill \mbox{$\text{EdgeContact}(B, x, y, w_o, h_o)$ for feasible $(x,y)$}
    \UNTIL{$\max_{(x,y)} R[y,x] \geq 0$}
    \STATE $(x^*, y^*) \gets \arg\max_{(x,y)} R[y,x]$ with random tie-breaking
    \STATE Place item $(w_o, h_o)$ at position $(x^*, y^*)$ in bin $B$
    \STATE $\mathcal{S} \gets \mathcal{S} \cup \{(w_o, h_o)\}$
    \STATE $\text{area} \gets \text{area} + w_o \cdot h_o$
    \STATE $\max_w \gets \max(\max_w, w_o)$, $\max_h \gets \max(\max_h, h_o)$
\ENDWHILE
\IF{$\text{area} = W \times H - 1$}
    \STATE $\mathcal{S} \gets \mathcal{S} \cup \{(1, 1)\}$
\ENDIF
\STATE Randomly shuffle $\mathcal{S}$
\RETURN $\mathcal{S}$
\end{algorithmic}
\end{algorithm}

% \textbf{100\% Set Construction:}
% \begin{enumerate}[label=(\roman*)]
%     \item Sample item dimensions from Gaussian-weighted probabilities (std $\sigma=2$)
%     \item Reject samples that exceed remaining bin capacity or dimension constraints
%     \item Place items at positions maximizing bottom-left heuristic score plus edge contact
%     \item Repeat until bin reaches 100\% utilization
%     \item Shuffle item ordering to simulate online arrival
% \end{enumerate}

\subsection{Training Results}

\begin{figure}[t]
    \centering

    \begin{minipage}{0.32\linewidth}
        \centering
        \includegraphics[width=1.0\linewidth]{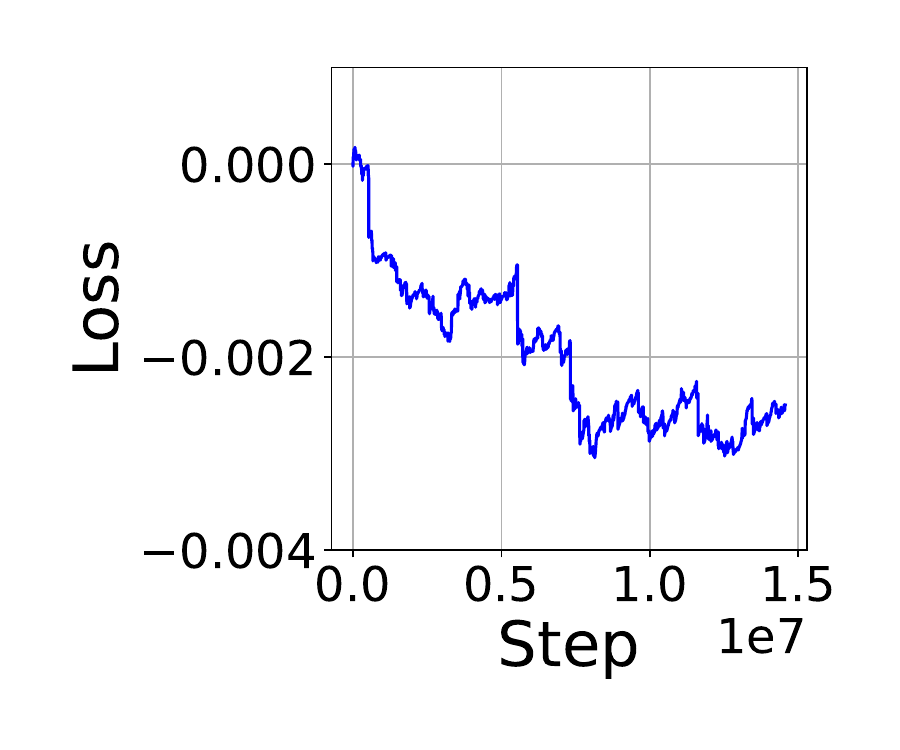}

        {\fontsize{8pt}{10pt}\selectfont (a) Policy Loss}
    \end{minipage}
    \hfill
    \begin{minipage}{0.32\linewidth}
        \centering
        \includegraphics[width=1.0\linewidth]{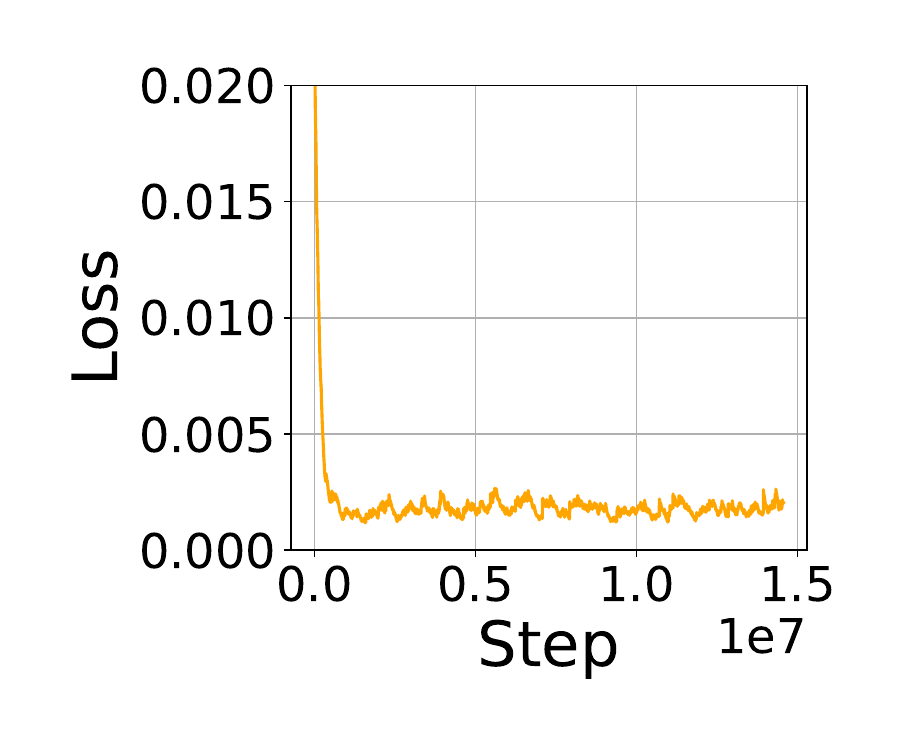}

        {\fontsize{8pt}{10pt}\selectfont (b) Value Loss}
    \end{minipage}
    \hfill
    \begin{minipage}{0.32\linewidth}
        \centering
        \includegraphics[width=1.0\linewidth]{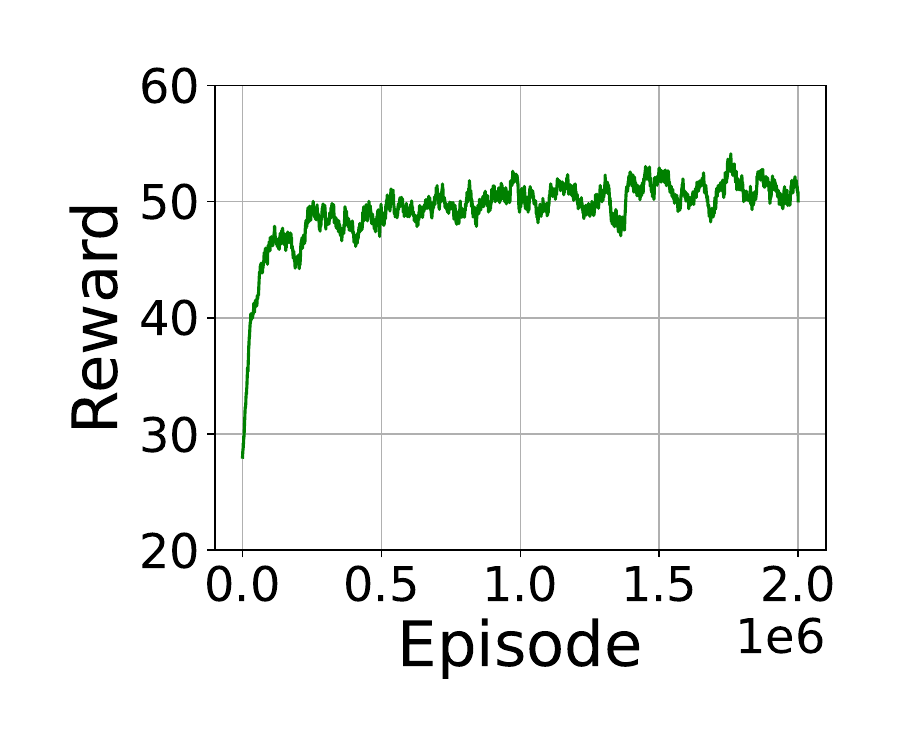}

        {\fontsize{8pt}{10pt}\selectfont (c) Reward}
    \end{minipage}

    \caption{Training curves for the proposed RL agent. (a) Policy loss, (b) value loss, and (c) reward.}
    \label{fig:training}
\end{figure}

Our A3C agent uses one global network and three parallel worker networks. Each worker independently interacts with its own environment and periodically updates the global network. This architecture enables stable learning by leveraging diverse experiences and reducing correlation among samples.
The agent is implemented in PyTorch~\cite{paszke2019pytorch} and trained for 7.2 million episodes (2.4 million per worker), requiring approximately 28 hours.
Fig.~\ref{fig:training} shows the training progress of the agent. The policy loss decreases over time, indicating that the agent improves its action selection. The value loss reflects the performance of the critic in predicting state values and also shows a downward trend, demonstrating stable learning of the value function.
The reward, obtained per episode, increases steadily throughout the training process, confirming the effectiveness of the learned policy in producing compact and efficient packing results.
These results validate that the A3C-based agent can learn a meaningful low-level policy capable of generalizing to diverse packing situations, serving as a reliable subroutine for the high-level search algorithm.

\begin{table*}[t]
\centering
\caption{Packing performance across different scenarios.}

% -------- Subtable (a) --------
{
\label{table:test1}
\centering
\begin{tabular}{ccccccccccccc}
\hline
\multicolumn{1}{|c|}{\textbf{Scenario}}              & \multicolumn{2}{c|}{S-R1A1}                             & \multicolumn{2}{c|}{S-R5A1}                             & \multicolumn{2}{c|}{S-R5A3}                             & \multicolumn{2}{c|}{D-R2A2O2}                           & \multicolumn{2}{c|}{D-R5A2O2}                           & \multicolumn{2}{c|}{D-R5A3O1}                           \\ \hline
\multicolumn{1}{|c|}{\textbf{Rotation}}              & \multicolumn{1}{c|}{No}    & \multicolumn{1}{c|}{Yes}   & \multicolumn{1}{c|}{No}    & \multicolumn{1}{c|}{Yes}   & \multicolumn{1}{c|}{No}    & \multicolumn{1}{c|}{Yes}   & \multicolumn{1}{c|}{No}    & \multicolumn{1}{c|}{Yes}   & \multicolumn{1}{c|}{No}    & \multicolumn{1}{c|}{Yes}   & \multicolumn{1}{c|}{No}    & \multicolumn{1}{c|}{Yes}   \\ \hline
\multicolumn{1}{|c|}{\textbf{Mean Utilization (\%)}} & \multicolumn{1}{c|}{78.88} & \multicolumn{1}{c|}{85.17} & \multicolumn{1}{c|}{86.45} & \multicolumn{1}{c|}{92.38} & \multicolumn{1}{c|}{94.77} & \multicolumn{1}{c|}{98.21} & \multicolumn{1}{c|}{85.94} & \multicolumn{1}{c|}{91.57} & \multicolumn{1}{c|}{91.99} & \multicolumn{1}{c|}{96.46} & \multicolumn{1}{c|}{95.86} & \multicolumn{1}{c|}{98.59} \\ \hline
\multicolumn{1}{|c|}{\textbf{Total Packed Items}}    & \multicolumn{1}{c|}{8715}  & \multicolumn{1}{c|}{9422}  & \multicolumn{1}{c|}{9559}  & \multicolumn{1}{c|}{10229} & \multicolumn{1}{c|}{10488} & \multicolumn{1}{c|}{10863} & \multicolumn{1}{c|}{9507}  & \multicolumn{1}{c|}{10127} & \multicolumn{1}{c|}{10179} & \multicolumn{1}{c|}{10661} & \multicolumn{1}{c|}{10602} & \multicolumn{1}{c|}{10905} \\ \hline
\multicolumn{1}{|c|}{\textbf{Total Packing Steps}}   & \multicolumn{6}{c|}{Same as total packed items}                                                                                                                             & \multicolumn{1}{c|}{5438}  & \multicolumn{1}{c|}{5706}  & \multicolumn{1}{c|}{6128}  & \multicolumn{1}{c|}{6333}  & \multicolumn{1}{c|}{5787}  & \multicolumn{1}{c|}{5893}  \\ \hline
\multicolumn{1}{l}{}                                 & \multicolumn{1}{l}{}       & \multicolumn{1}{l}{}       & \multicolumn{1}{l}{}       & \multicolumn{1}{l}{}       & \multicolumn{1}{l}{}       & \multicolumn{1}{l}{}       & \multicolumn{1}{l}{}       & \multicolumn{1}{l}{}       & \multicolumn{1}{l}{}       & \multicolumn{1}{l}{}       & \multicolumn{1}{l}{}       & \multicolumn{1}{l}{}      
\end{tabular}
}

\vspace{-6pt}
{\fontsize{8pt}{10pt}\selectfont (a) Performance on \textit{Random} instances}

\vspace{8pt}

% \vspace{-5pt}
% -------- Subtable (b) --------
{
\label{table:test2}
\centering
\begin{tabular}{|c|cccccc|cc|cc|cc|}
\hline
\textbf{Scenario}                                                                             & \multicolumn{2}{c|}{S-R1A1}                            & \multicolumn{2}{c|}{S-R5A1}                            & \multicolumn{2}{c|}{S-R5A3}       & \multicolumn{2}{c|}{D-R2A2O2}     & \multicolumn{2}{c|}{D-R5A2O2}     & \multicolumn{2}{c|}{D-R5A3O1}     \\ \hline
\textbf{Repacking}                                                                & \multicolumn{1}{c|}{No}    & \multicolumn{1}{c|}{Yes}  & \multicolumn{1}{c|}{No}    & \multicolumn{1}{c|}{Yes}  & \multicolumn{1}{c|}{No}    & Yes  & \multicolumn{1}{c|}{No}    & Yes  & \multicolumn{1}{c|}{No}    & Yes  & \multicolumn{1}{c|}{No}    & Yes  \\ \hline
\textbf{Mean Utilization (\%)}                                                                & \multicolumn{1}{c|}{81.52} & \multicolumn{1}{c|}{100}  & \multicolumn{1}{c|}{95.54} & \multicolumn{1}{c|}{100}  & \multicolumn{1}{c|}{97.67} & 100  & \multicolumn{1}{c|}{86.98} & 100  & \multicolumn{1}{c|}{96.49} & 100  & \multicolumn{1}{c|}{96.53} & 100  \\ \hline
\textbf{Num. of Full Bins}                                                                    & \multicolumn{1}{c|}{684}   & \multicolumn{1}{c|}{3000} & \multicolumn{1}{c|}{2213}  & \multicolumn{1}{c|}{3000} & \multicolumn{1}{c|}{2395}  & 3000 & \multicolumn{1}{c|}{865}   & 3000 & \multicolumn{1}{c|}{1999}  & 3000 & \multicolumn{1}{c|}{2240}  & 3000 \\ \hline
\textbf{\begin{tabular}[c]{@{}c@{}}Mean Num. of\\ Packed Items\end{tabular}}                  & \multicolumn{1}{c|}{6.85}  & \multicolumn{1}{c|}{8.17} & \multicolumn{1}{c|}{7.85}  & \multicolumn{1}{c|}{8.17} & \multicolumn{1}{c|}{7.97}  & 8.17 & \multicolumn{1}{c|}{7.44}  & 8.17 & \multicolumn{1}{c|}{7.84}  & 8.17 & \multicolumn{1}{c|}{7.91}  & 8.17 \\ \hline
\textbf{\begin{tabular}[c]{@{}c@{}}Mean Num. of\\ Packing Steps\end{tabular}}                 & \multicolumn{6}{c|}{Same as mean of packed items}                                                                                                   & \multicolumn{1}{c|}{4.16}  & 4.34 & \multicolumn{1}{c|}{4.40}  & 4.34 & \multicolumn{1}{c|}{4.23}  & 4.34 \\ \hline
\textbf{\begin{tabular}[c]{@{}c@{}}Mean Num. of\\ Repacking Items\end{tabular}}               & \multicolumn{1}{c|}{-}     & \multicolumn{1}{c|}{2.42} & \multicolumn{1}{c|}{-}     & \multicolumn{1}{c|}{0.39} & \multicolumn{1}{c|}{-}     & 0.30 & \multicolumn{1}{c|}{-}     & 1.90 & \multicolumn{1}{c|}{-}     & 0.52 & \multicolumn{1}{c|}{-}     & 0.39 \\ \hline
\textbf{\begin{tabular}[c]{@{}c@{}}Mean Num. of Steps\\ including Repacking\end{tabular}}     & \multicolumn{1}{c|}{-}     & \multicolumn{1}{c|}{1.00} & \multicolumn{1}{c|}{-}     & \multicolumn{1}{c|}{0.27} & \multicolumn{1}{c|}{-}     & 0.21 & \multicolumn{1}{c|}{-}     & 0.80 & \multicolumn{1}{c|}{-}     & 0.34 & \multicolumn{1}{c|}{-}     & 0.26 \\ \hline
\textbf{\begin{tabular}[c]{@{}c@{}}Mean Taken Time (s)\\ for Repacking Solution\end{tabular}} & \multicolumn{1}{c|}{-}     & \multicolumn{1}{c|}{2.55} & \multicolumn{1}{c|}{-}     & \multicolumn{1}{c|}{1.43} & \multicolumn{1}{c|}{-}     & 0.90 & \multicolumn{1}{c|}{-}     & 2.82 & \multicolumn{1}{c|}{-}     & 1.33 & \multicolumn{1}{c|}{-}     & 0.79 \\ \hline
\end{tabular}
}

\vspace{3pt}
{\fontsize{8pt}{10pt}\selectfont (b) Performance on \textit{100\% set} with and without repacking}

% \vspace{-20pt}
\end{table*}

% We define eight distinct scenarios to evaluate our bin packing framework under varying robotic configurations and perception constraints, as summarized in Table~\ref{table:scenario}. Each scenario is denoted by a label that encodes:
% \begin{itemize}
%     \item \textbf{S} (single robot) or \textbf{D} (dual robot): the robot configuration,
%     \item \textbf{R}$x$: the maximum number of items recognized by the vision system,
%     \item \textbf{A}$y$: the number of items accessible per robot,
%     \item \textbf{O}$z$: the number of items located in the overlapping workspace for dual-robot settings.
% \end{itemize}
% For example, Scenarios S-R1A1, S-R5A1, S-R5A3, and S-R5A5 correspond to single manipulator configurations,
% while D-R2A2O2, D-R5A2O2, D-R5A3O1, and D-R5A5O5 represent dual manipulator configurations.
% Scenario S-R1A1 represents a minimal configuration where one robot can recognize and access only a single object.
% In contrast, Scenario D-R5A5O5 reflects the most capable setting, where two robots can recognize five objects, each access five objects, and share access to all five through an overlapping workspace.

\subsection{Test Environment}
\subsubsection{Scenarios}

% We define six distinct scenarios to evaluate our bin packing framework under varying robotic configurations and perception constraints, as summarized in Table~\ref{table:scenario} and shown in Fig.~\ref{fig:scenarios}.
% Each scenario label encodes the following:
% \begin{itemize}
%     \item \textbf{S} (single robot) or \textbf{D} (dual robot): robot configuration,
%     \item \textbf{R}: maximum number of recognized items,
%     \item \textbf{A}: number of accessible items per robot,
%     \item \textbf{O}: number of items in the overlapping workspace (for dual-robot settings only).
% \end{itemize}

% Scenarios S-R1A1, S-R5A1, and S-R5A3 represent single-arm configurations,
% while D-R2A2O2, D-R5A2O2, and D-R5A3O1 correspond to dual-arm settings.
% S-R1A1 denotes a minimal setup where one robot can recognize and access only a single item.
% In contrast, D-R5A3O1 reflects the most capable configuration, where two robots can recognize five items, each access three, and share one item in the overlapping region.

We define six distinct scenarios to evaluate the proposed bin packing framework under varying robotic configurations and perception constraints, as summarized in Table~\ref{table:scenario} and illustrated in Fig.~\ref{fig:scenarios}.
Each scenario label is encoded as follows:
\begin{itemize}
    \item \textbf{S} (single robot) or \textbf{D} (dual robot): robot configuration,
    \item \textbf{R}: maximum number of recognized items,
    \item \textbf{A}: maximum number of accessible items per robot,
    \item \textbf{O}: maximum number of items in the overlapping workspace (for dual-robot settings only).
\end{itemize}
Scenarios S-R1A1, S-R5A1, and S-R5A3 represent single-robot configurations,
while D-R2A2O2, D-R5A2O2, and D-R5A3O1 correspond to dual-robot settings.
S-R1A1 denotes the minimal case, where a single robot can recognize and access only one item at a time.
In contrast, D-R5A3O1 represents the most capable configuration, where two robots can recognize five items, each can access three items, and one item lies within the overlapping workspace shared by both robots.
The maximum number of items that can be accessed by all robots, denoted as $n_{\max}$, is given by $n_{\max} = n_{\text{A}} \times n_{\text{robot}} - n_{\text{O}} \times (n_{\text{robot}} - 1)$,
where $n_{\text{A}}$ denotes the number of accessible items per robot, 
$n_{\text{O}}$ represents the number of items located in the overlapping workspace, 
and $n_{\max} \leq n_{\text{R}}$, with $n_{\text{R}}$ being the number of recognized items.

\subsubsection{Conveyor Logic}
Since our search algorithm considers both robot accessibility and task allocation, 
the conveyor belt logic for positioning incoming items is a critical component of the experimental setup. 
For single-robot scenarios or dual-robot configurations where both robots share the same accessible region (e.g., $n_\text{A} = n_\text{O}$ in D-R2A2O2 and D-R5A2O2), 
items can be continuously positioned along the conveyor. 
However, in configurations such as D-R5A3O1, where the maximum number of items accessible to each robot ($n_\text{A}$) differs from that in the overlapping region ($n_\text{O}$),
exclusive zones arise that can be reached by only one robot. 
If more items accumulate in one exclusive zone, 
the advantages of using dual robots diminish due to workload imbalance.
To maintain balanced item allocation, we model the conveyor control logic at each time step as follows.

\textbf{Notation:}
\begin{itemize}
    \item $k_{\text{total}}$: Total number of items on the conveyor
    \item $k_{\alpha}$: Number of items in the exclusive region of the rear robot
    \item $k_{\beta}$: Number of items in the exclusive region of the front robot
    \item $k_{\text{O}}$: Number of items in the overlapping region (accessible to both robots)
    \item $t$: Time step before conveyor operation
    \item $t+1$: Time step after conveyor operation
\end{itemize}
When $k_{\text{total},t+1} \geq n_{\text{max}}$, 
where $n_{\text{max}} = 2n_\text{A} - n_\text{O}$ denotes the maximum number of items that both robots can collectively access, 
items are continuously positioned along the conveyor as in the fully shared-region cases. 
When $k_{\text{total},t+1} < n_{\text{max}}$, 
the item distribution across regions is determined as follows:
\begin{align}
k_{\text{O},t+1} &= \min\!\left(k_{\text{total},{t+1}} - k_{\alpha,t}, \, n_{\text{O}}\right), \\
k_{\alpha,{t+1}} &= \max\!\left(k_{\alpha,t}, \, \left\lfloor\frac{k_{\text{total},t+1}}{2}\right\rfloor + 1 - k_{\text{O},t+1}\right), \\
k_{\beta,{t+1}} &= k_{\text{total},t+1} - k_{\alpha,t+1} - k_{\text{O},t+1}.
\end{align}

By controlling the conveyor belt to place items sequentially into the $\alpha$, $\text{O}$, and $\beta$ regions according to the computed quantities, 
the system maintains a balanced allocation of items between the workspaces of the two robots 
even when their accessible regions differ, thereby preserving the efficiency gains of dual-robot operation.
All following tests are conducted under the assumption that the conveyor belt operates according to the described logic at every time step.

\subsection{Algorithm Tests}
\label{sec:Algorithm_Tests}

Before discussing each result in detail, we briefly summarize the purpose and scope of the ablation studies conducted to verify the contribution of each component in the proposed hierarchical framework.
In summary, the ablation results demonstrate that:
\begin{itemize}
\item under the consideration of rotation and multiple perceived items, the tree search component improves orientation selection and packing efficiency;
\item the dual-manipulator setting significantly reduces packing steps while maintaining high utilization;
\item the tree search handles multiple reachable items and performs task allocation, especially in dual-manipulator settings;
\item the repacking algorithm further enhances bin utilization across all scenarios within a limited planning time;
\item the reinforcement learning module alone achieves higher utilization than the baseline algorithms.
\end{itemize}
Detailed analyses of each case are presented in the following subsections.

\subsubsection{Test with Random Instances}

% We evaluate our method on the \textit{Random} instances with 1000 episodes per scenario, where each episode involves packing a single bin.
% All scenarios share the same item arrival sequence across the entire test.
% We first evaluate our method without repacking across eight scenarios, with and without item rotation. As shown in Table~\ref{table:test1}, both bin utilization and the total number of packed items consistently improve when rotation is enabled. For instance, in Scenario S-R1A1, the utilization increases from 78.88\% to 85.17\%, and the number of packed items from 8715 to 9422. Similar trends are observed across all scenarios, with Scenario S-R5A5 achieving the highest utilization of 98.84\%.
% We then test repacking in four scenarios where there is room for improvement: S-R1A1, S-R5A1, D-R2A2O2, and D-R5A2O2. \bl{Fig.~\ref{fig:test1_repacking} shows that even a short additional planning time for repacking solution significantly improves the final bin utilization. In Scenario S-R1A1, the utilization increases from 85.03\% to 97.97\%, and in D-R2A2O2 from 91.49\% to 99.36\%.
% Scenarios with more recognized items per step (e.g., S-R5A5, D-R5A5O5) require more time to expand the search tree, resulting in fewer updates within the time budget. Nevertheless, the final utilization remains comparable to that of other scenarios. These settings also result in fewer repacking actions, which may help reduce motion overhead during execution.}

We evaluate our method on the \textit{Random} instances with 1000 episodes per scenario, where each episode involves packing a single bin.
In the \textit{Random} instance, the items that has not been packed at the current episode, are not discarded but are reconsidered in subsequent planning stages. 
Such items are revisited either during the repacking phase or when planning for the next bin, ensuring that all items have the opportunity to be packed. 
This cumulative process over 1,000 bins is reflected in the total number of packed items reported in Table~\ref{table:test1}(a).
All scenarios share the same item arrival sequence across the entire test.

We first evaluate our method without repacking across six scenarios, with and without item rotation.
When performing packing only, the \textsc{Selection} described in Alg.~\ref{alg:pack} (line~\ref{line:2-18}) is skipped, and the total planning time for all scenarios averages less than 1.2 seconds.
As shown in Table~\ref{table:test1}(a), both bin utilization and the total number of packed items consistently improve when rotation is enabled.
For instance, in Scenario S-R1A1, the utilization increases from 78.88\% to 85.17\%, and the number of packed items from 8715 to 9422, achieving higher throughput within the same number of bins. Similar trends are observed across all scenarios, with Scenario D-R5A3O1 achieving the highest utilization of 98.59\%.
Furthermore, the influence of the number of perceived items on performance is verified by comparing S-R1A1 vs.~S-R5A1 and D-R2A2O2 vs.~D-R5A2O2. 
In addition, when comparing S-R5A1 vs.~S-R5A3 and D-R5A2O2 vs.~D-R5A3O1, where each robot has more reachable objects, the results show consistent improvements in utilization. 
In the dual-arm cases (D-R5A2O2 vs.~D-R5A3O1), even with a larger number of packed items, the total number of packing steps decreased. 
These results collectively verify that the proposed candidate-selection strategy is valid and that the method successfully performs task allocation between the two manipulators, directly indicating improved throughput per cell.

\begin{figure}[t]
    \centering

    \begin{minipage}{0.48\linewidth}
        \centering
        \includegraphics[width=1.0\linewidth]{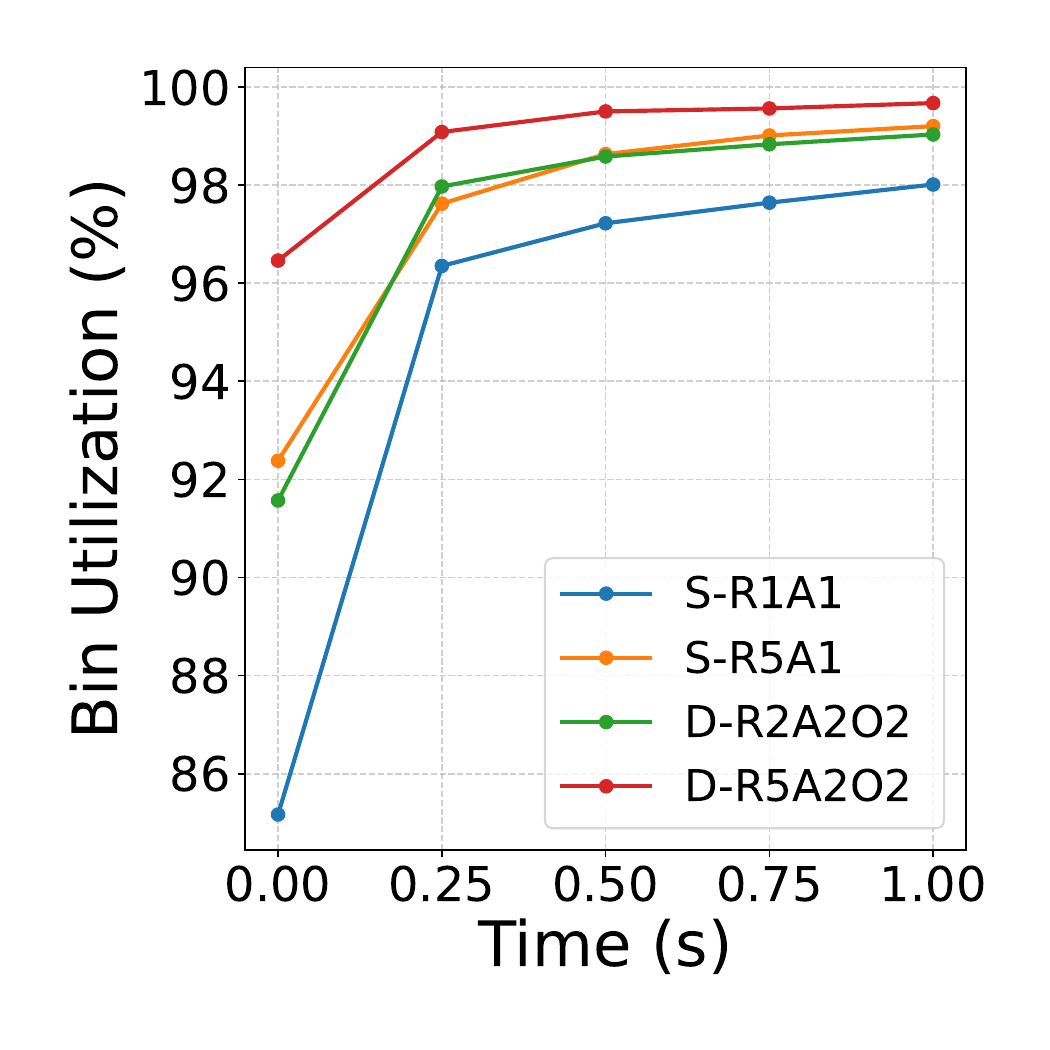}

        {\fontsize{8pt}{10pt}\selectfont (a) Bin utilization over time}
        \label{fig:uti_with_repacking}
    \end{minipage}
    \hfill
    \begin{minipage}{0.48\linewidth}
        \centering
        \includegraphics[width=1.0\linewidth]{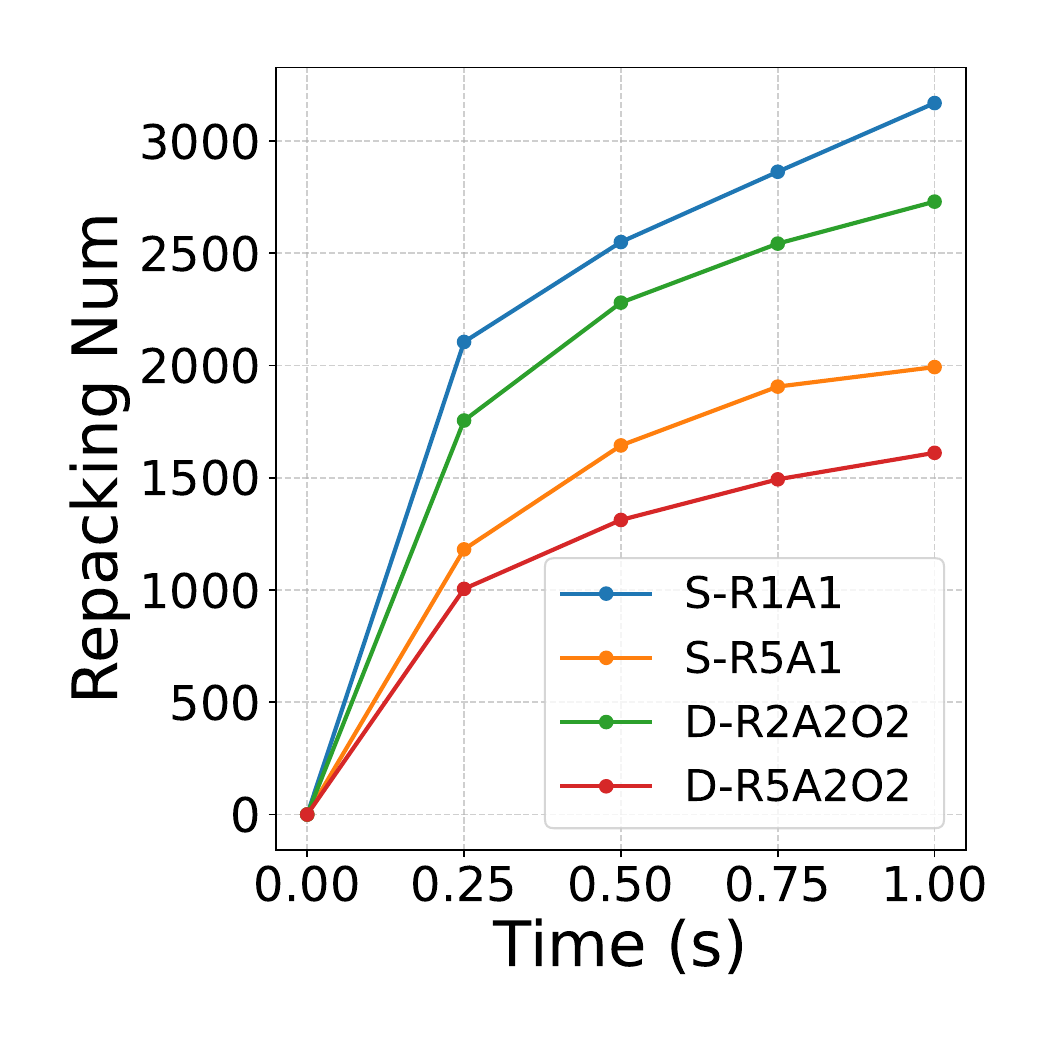}

        {\fontsize{8pt}{10pt}\selectfont (b) Total number of repacked items}
        \label{fig:num_with_repacking}
    \end{minipage}

    \caption{
    Impact of repacking planning time on packing performance.
    (a) Bin utilization improves as more planning time is allocated. 
    (b) The total number of repacked items increases with longer planning time.}
    \label{fig:test1_repacking}
\end{figure}

We then test repacking in four scenarios where there is room for improvement: S-R1A1, S-R5A1, D-R2A2O2, and D-R5A2O2.
In these repacking experiments, the number of candidates selected by the \textsc{Selection} function is fixed to two to compare the utilization achieved under a limited additional planning time for repacking.
Fig.~\ref{fig:test1_repacking} shows that even a short planning time of 1 second for repacking significantly improves the final bin utilization.
In Scenario S-R1A1, the utilization increases from 85.17\% to 98.01\%, and in D-R2A2O2 from 91.57\% to 99.03\%.
In Scenario S-R5A1, the utilization improves from 92.38\% to 99.20\%, and in D-R5A2O2 from 96.46\% to 99.67\%.
The fact that such high improvements are achieved within a short planning time indicates that the proposed last-packed-first search strategy efficiently identifies promising repacking policies.
Scenarios with more known items result in fewer repacking actions, which helps reduce motion overhead during execution.

\begin{figure}[t]
    \centering

    \includegraphics[width=0.9\linewidth]{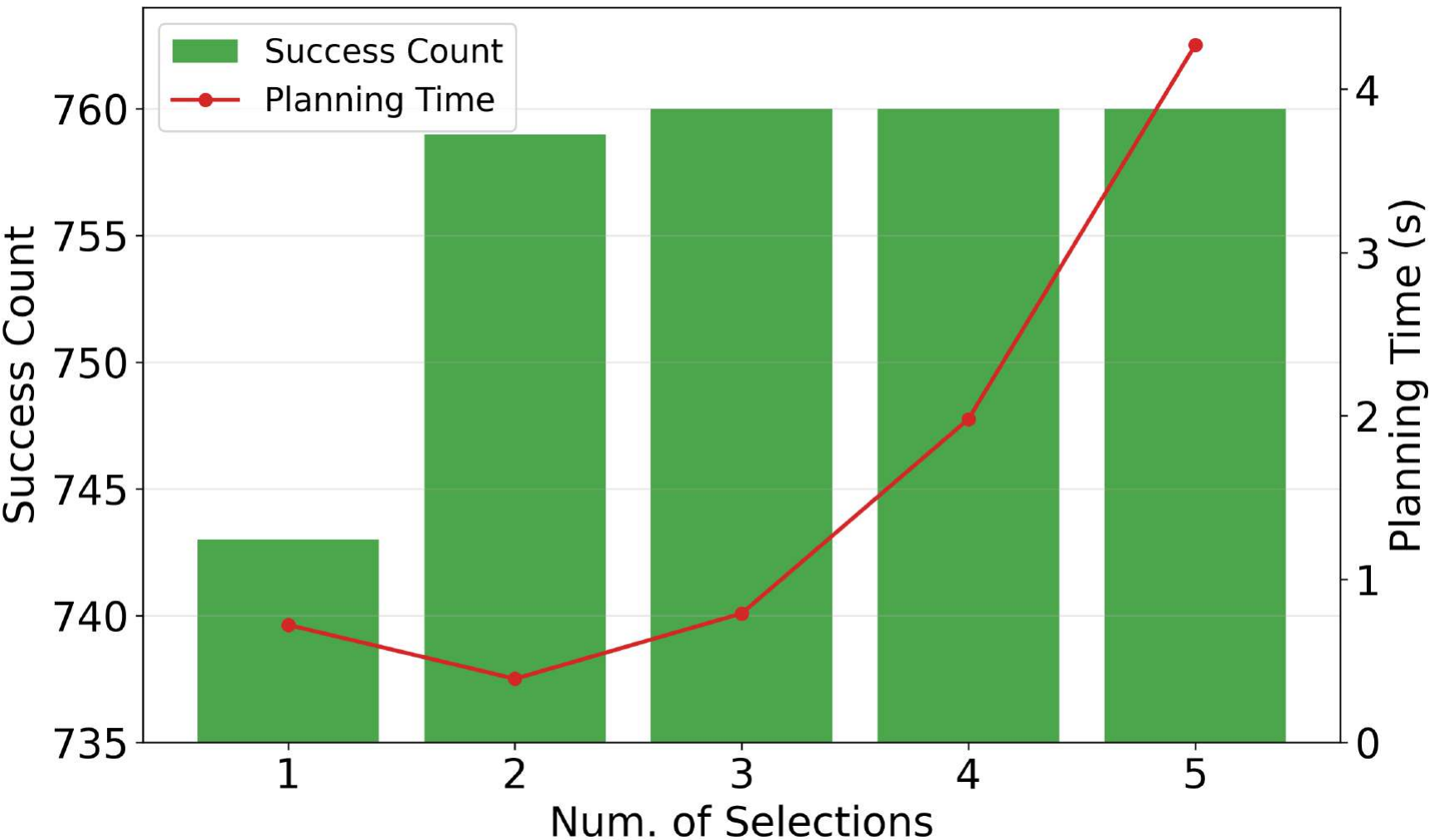}

    \caption{
    Repacking success count and planning time with varying numbers of candidates selected by the \textsc{Selection} in D-R5A3O1 on the \textit{100\% set}.
    }
    \label{fig:sensitibity_tests}
\end{figure}

\subsubsection{Test with 100\% set}
\label{sec:repacking}

We evaluate our method on the \textit{100\% set} with 3000 episodes, where each episode uses the same fixed item arrival sequence per scenario.
In the \textit{100\% set} experiments, items that have no valid placement are re-evaluated during the repacking phase, and if infeasible placements remain, those items are excluded from further consideration, since the item sets are predefined for each bin.

When performing packing only, the \textsc{Selection} in Alg.~\ref{alg:pack} (line~\ref{line:2-18}) is skipped, and the total planning time for all scenarios averages less than 1.3~seconds.
As shown in Table~\ref{table:test2}(b), our method achieves high utilization even without repacking.
Furthermore, performance improves as the number of perceived and reachable items increases, confirming that the hierarchical decision process leverages additional information.

In addition, we conduct an experiment on the \textit{100\% set} to measure the repacking success count and computation time with varying numbers of candidates selected by the \textsc{Selection} function in D-R5A3O1, as presented in Fig.~\ref{fig:sensitibity_tests}.
The \textsc{Selection} process is applied until the number of remaining items in the search tree drops below three; when fewer than three items remains, pruning is skipped to preserve remaining candidates, since the search space becomes sufficiently small due to additional early-termination conditions implemented in Alg.2 (lines~10–11,29–30).
When repacking is enabled with this setting, 100\% utilization is achieved across all scenarios.
The number of fully packed bins increases significantly with repacking, from 684 to 3000 in S-R1A1 and from 865 to 3000 in D-R2A2O2.
Repacking improves packing quality at the cost of moderate computation time, with an average of less than 3 seconds to find a repacking solution.
These findings further confirm that the proposed hierarchical framework achieves near-optimal utilization under realistic time constraints.

\subsubsection{Compare with Baseline}

\begin{table}[htbp]
\centering
\caption{Comparison of bin utilization (\%) with baseline methods in $4 \times 4$ and $5 \times 5$ bin environments.}
\label{table:compare}

\begin{minipage}{0.48\linewidth}
\centering
\begin{tabular}{cc}
\hline
\multicolumn{1}{|c|}{Method}        & \multicolumn{1}{c|}{Uti. (\%)}      \\ \hline
\multicolumn{1}{|c|}{Deep-Pack}     & \multicolumn{1}{c|}{94.87}          \\ \hline
\multicolumn{1}{|c|}{BERM}          & \multicolumn{1}{c|}{97.23}          \\ \hline
\multicolumn{1}{|c|}{\textbf{Ours}} & \multicolumn{1}{c|}{\textbf{98.23}} \\ \hline
\multicolumn{1}{l}{}                & \multicolumn{1}{l}{}                \\
\multicolumn{1}{l}{}                & \multicolumn{1}{l}{}            
\end{tabular}
\vspace{5pt}

{\fontsize{8pt}{10pt}\selectfont (a) $4 \times 4$ bin}
\end{minipage}
\hfill
\begin{minipage}{0.48\linewidth}
\centering
\begin{tabular}{|c|c|}
\hline
Method & Uti. (\%) \\ \hline
ShelfNextFit & 75.80 \\ \hline
Skyline & 85.33 \\ \hline
Deep-Pack & 91.48 \\ \hline
BERM & 94.30 \\ \hline
\textbf{Ours} & \textbf{96.58} \\ \hline
\end{tabular}
\vspace{5pt}

{\fontsize{8pt}{10pt}\selectfont (b) $5 \times 5$ bin}
\end{minipage}

\end{table}

To validate the effectiveness of our method, we compare it with existing baselines, including heuristic methods (Skyline~\cite{Skyline}, ShelfNextFit~\cite{ShelfNextFit}) and learning-based approaches (Deep-Pack~\cite{deeppack}, BERM~\cite{brain-inspired}).
The evaluation follows the same setup as the baselines in $4 \times 4$ and $5 \times 5$ bin environments under a fully online setting without rotation or repacking, ensuring fair comparison.
Table~\ref{table:compare} shows that our method achieves the highest mean bin utilization in both settings—98.23\% in the $4 \times 4$ bin (compared to 94.87\% for Deep-Pack and 97.23\% for BERM), and 96.58\% in the $5 \times 5$ bin (compared to 91.48\% and 94.30\%, respectively).
Assuming that each pallet or tote corresponds to a fixed container volume, this improvement directly translates to a reduction of roughly one pallet for every 20–40 bins packed, resulting in measurable logistics cost savings and reduced storage requirements.
The superior performance of our method can be attributed to the DRL formulation. 
Compared with heuristic-based baselines~\cite{ShelfNextFit,Skyline}, which usually follow sequential placement heuristics, our RL agent is free to explore and select placements from any direction within the bin, allowing more flexible decision-making.
Furthermore, our method surpasses existing DRL-based algorithms~\cite{deeppack,brain-inspired} due to two main design choices:
\begin{itemize}
\item the use of diverse training datasets that improve generalization across different item distributions,
\item a reward formulation defined as the number of occupied cells adjacent to the packed item (described in Sec.~\ref{sec:low}), which provides dense and flexible rewards and induces generalizable packing behaviors. 
\end{itemize}

\subsection{Simulation Tests}
\subsubsection{Configuration of Simulation Environment}
\begin{figure}[!t]
    \centering
    \includegraphics[width=0.9\linewidth]{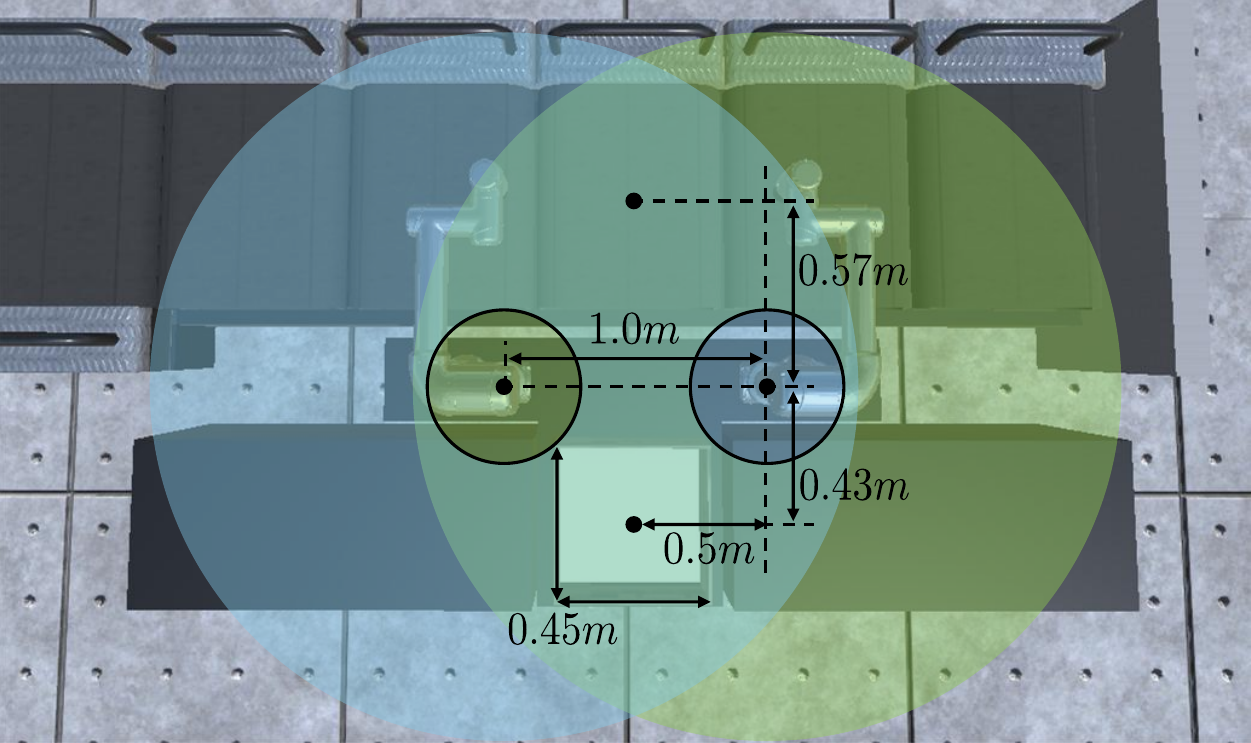}
    \vspace{-3pt}
    
    {\fontsize{8pt}{10pt}\selectfont (a) Layout of the dual-arm simulation environment (D-R5A3O1)}

    \vspace{8pt} % 두 그림 사이 간격 조정
    \includegraphics[width=0.9\linewidth]{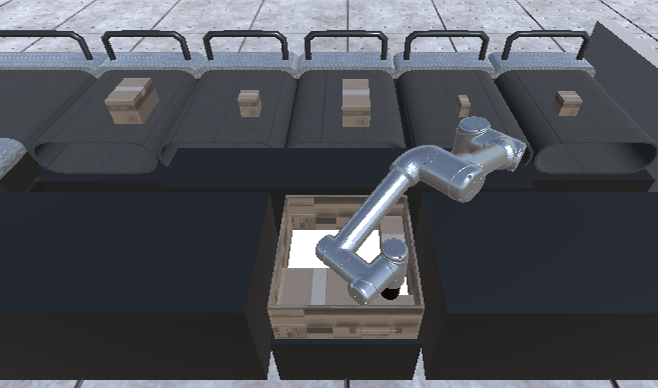}
    \vspace{-3pt}
    
    {\fontsize{8pt}{10pt}\selectfont (b) Single-arm packing scene (S-R5A3)}

    \vspace{8pt} % 두 그림 사이 간격 조정

    \includegraphics[width=0.9\linewidth]{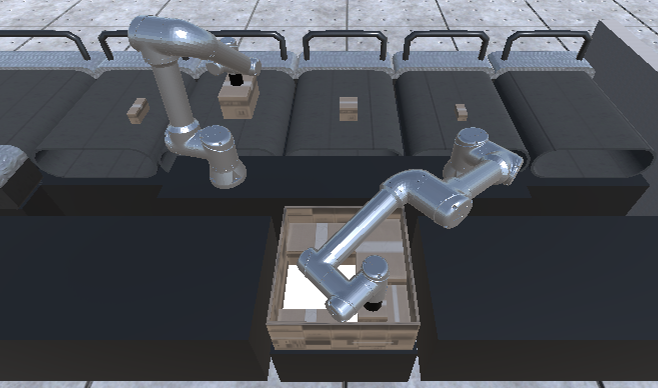}
    \vspace{-3pt}
    
    {\fontsize{8pt}{10pt}\selectfont (c) Dual-arm packing scene (D-R5A3O1)}
    \vspace{-6pt}

    \caption{Overview of the simulation environment and packing scenes. In (a), the layout of the dual-arm simulation environment is shown, where the circles indicate the workspace of each manipulator, while (b) and (c) show the packing scenes of the single-arm and dual-arm configurations, respectively.}
    \label{fig:simulation_env}
\end{figure}

The simulation environment is developed using Unity~\cite{unity_robotics_hub} to evaluate the proposed framework under physical constraints. 
Universal Robots UR5e is used as the manipulator in both single- and dual-arm configurations.
A square bin of 45~cm~$\times$~45~cm is used, discretized into a $10~\times~10$ grid where each cell corresponds to 4.5~cm.
The simulation environment layout for the dual-robot scenario, and the packing scenes of single-arm and dual-arm configurations are illustrated in Fig.~\ref{fig:simulation_env}.
The design of the simulation environment is primarily guided by the workspaces of both manipulators.
Each robot is positioned such that both arms can fully cover the bin area while maintaining sufficient reach to grasp incoming items on the conveyor belt.
To prevent mutual motion interference, the robots are spaced approximately half of their reach distance apart, which reflects safety and accessibility considerations.
For the single-robot configuration, only the right arm is used to ensure consistent comparison under identical environmental conditions.

Fig.~\ref{fig:Makespan} illustrates the action sequence of the robot during the packing process.
The interval between incoming items on the conveyor belt is set to 0.7~m, and the conveyor speed is configured to 0.5~m/s.
When the conveyor operates and a new item arrives, the planning module generates a placement plan, after which the robot initiates the packing task.
The packing operation consists of repeated pick-and-place actions.
Once the picking of the current item on the conveyor is completed, the conveyor operates again to transport the next item, thereby allowing conveyor operation and robotic placement to proceed in parallel.
Once the new item has arrived and the planning for the next step is completed, followed by the completion of the previous placement, the next operation begins (if repacking is required, the repacking phase is executed first).

\subsubsection{Makespan Evaluation}
\begin{figure*}[t]
    \centering

    \includegraphics[width=1.0\linewidth]{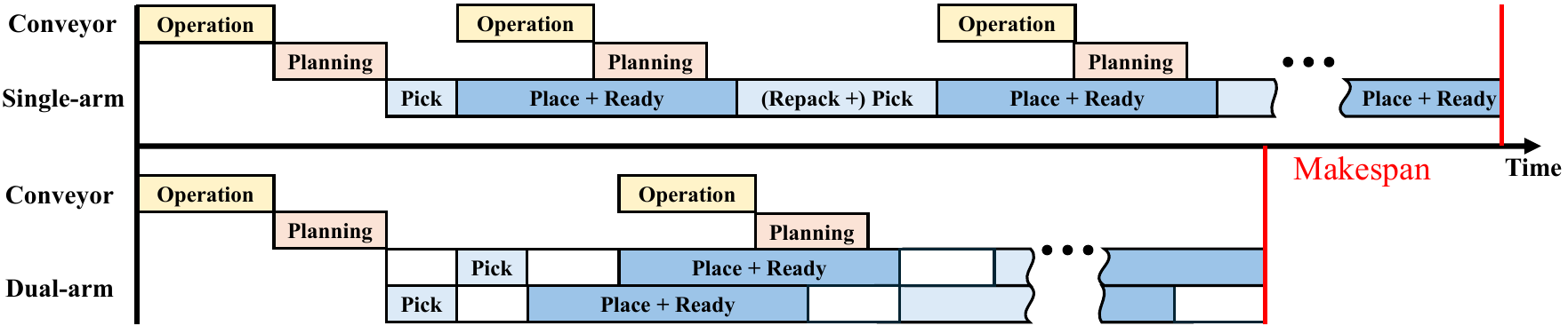}

    \caption{Action sequence and makespan of single-arm and dual-arm configurations.}
    \label{fig:Makespan}
\end{figure*}

In this paper, we define the makespan as the total time required for the robot(s) to complete the loading process within a single episode, including conveyor operation, planning, and execution, as shown in Fig.~\ref{fig:Makespan}.
For motion planning, cuRobo~\cite{curobo_report23} is employed in both configurations, while Stop-N-Go~\cite{stopngo} is integrated to coordinate dual-arm execution. 
Each setup is tested over 30 episodes using the \textit{100\% set} instances. The average time between the conveyor activation and the arrival of a new item is approximately 1.7 seconds.

As shown in Table~\ref{table:simulation}, the dual-arm scenario achieves a shorter makespan than the single-arm scenario, despite requiring longer per-step planning durations. 
Specifically, it reduces the overall makespan by approximately 24\% in both repacking and non-repacking cases, corresponding to a proportional increase in cell throughput under identical task conditions.
Since planning and execution proceed concurrently, planning for the next step is typically completed during the current execution, thereby minimizing idle time. 
Although repacking introduces slight overhead in planning and execution, it improves bin utilization (as discussed in Sec.~\ref{sec:repacking}), which justifies the additional cost. 
In instances with an odd number of items, the final item is always handled by a single robot, limiting the relative gain in makespan reduction. 
These results demonstrate that the proposed framework can generate physically feasible and time-efficient plans for dual-arm robotic systems.

\subsection{Physical Robot Test}

\begin{figure}[!t]
    \centering
    \includegraphics[width=0.5\linewidth]{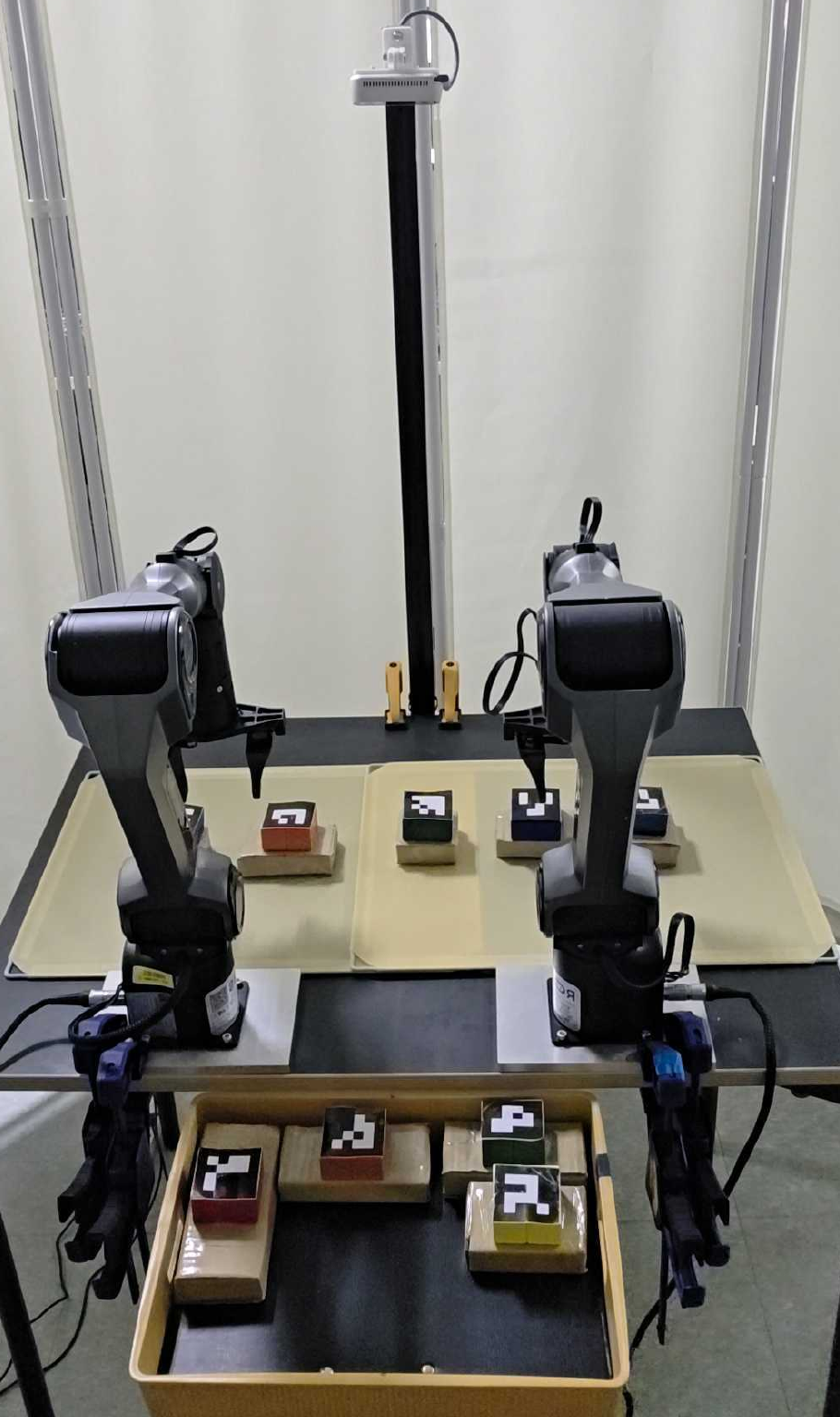}

    \caption{Real dual-arm packing experiment setup showing the manipulators, bin, and vision-based detection system.}
    \label{fig:real_robot}
\end{figure}

In the physical robot experiments, a 32~cm~$\times$~26~cm bin is used with 1~cm margins on both sides, corresponding to a 3~cm grid resolution.
Two AgileX Piper manipulators~\cite{agilex_piper} are employed for dual-arm operation, following the D-R5A3O1 scenario setting.
ArUco markers are attached to the objects for vision-based detection and pose estimation.
The experimental setup and dual-arm operation are illustrated in Fig.~\ref{fig:real_robot}.
Motion planning for dual manipulators is performed using cuRobo~\cite{curobo_report23} and Stop-N-Go~\cite{stopngo}.
Through these experiments, we demonstrated the feasibility of applying the proposed framework to real dual-arm packing tasks in a logistics environment.

% \begin{figure}[!t]
%     \centering
%     \includegraphics[width=0.9\linewidth]{figure/layout.pdf}
%     \vspace{-3pt}
    
%     {\fontsize{8pt}{10pt}\selectfont (a) Layout of the dual-arm simulation environment (scenario D-R5A3O1). The circles indicate the workspace of each manipulator.}

%     \vspace{8pt} % 두 그림 사이 간격 조정

%     \includegraphics[width=0.9\linewidth]{figure/Makespan.pdf}
%     \vspace{-3pt}
    
%     {\fontsize{8pt}{10pt}\selectfont (b) Dual-arm packing scene (D-R5A3O1)}
%     \vspace{-6pt}

%     \caption{Configuration of the simulation environment.}
%     \label{fig:sim_config}
% \end{figure}

\begin{table}
\centering
\caption{Planning and Execution Time in Single- and Dual-Arms.}
\label{table:simulation}
\begin{tabular}{|c|cc|cc|}
\hline
\textbf{Scenario}                                                                   & \multicolumn{2}{c|}{S-R5A3}          & \multicolumn{2}{c|}{D-R5A3O1}       \\ \hline
\textbf{Repacking}                                                                  & \multicolumn{1}{c|}{No}     & Yes    & \multicolumn{1}{c|}{No}    & Yes    \\ \hline
\textbf{Task Planning Time (s)}                                                     & \multicolumn{1}{c|}{0.71}   & 0.76   & \multicolumn{1}{c|}{0.77}  & 0.85   \\ \hline
\textbf{Motion Planning Time (s)}                                                   & \multicolumn{1}{c|}{0.73}   & 0.77   & \multicolumn{1}{c|}{1.46}  & 1.60    \\ \hline
\textbf{Execution Time (s)}                                                         & \multicolumn{1}{c|}{15.50}   & 15.93  & \multicolumn{1}{c|}{21.22} & 22.39  \\ \hline
\textbf{\begin{tabular}[c]{@{}c@{}}Average Makespan\\ per episode (s)\end{tabular}} & \multicolumn{1}{c|}{124.48} & 130.98 & \multicolumn{1}{c|}{94.29} & 100.12 \\ \hline
\end{tabular}
% \vspace{-20pt}
\end{table}

\section{Conclusion}

This paper presented a hierarchical bin packing framework that combines heuristic tree search and deep reinforcement learning to solve 2D bin packing problems. The method supports repacking and dual-arm coordination through integrated task and motion planning. A low-level RL agent proposes placement positions, while the high-level search explores packing and unpacking strategies under item availability and accessibility constraints. Experiments demonstrate near-optimal bin utilization and show that the framework generates physically feasible, time-efficient plans, reducing execution time in dual-arm scenarios. These results highlight the benefits of repacking and parallel execution for practical robotic systems.
Future work is left for extending the current 2D framework to 3D by replacing the binary grid matrix representation with a height-map formulation, while incorporating dual-arm workspace constraints for three-dimensional packing tasks. % review 1-1

\bibliographystyle{IEEEtran}
\bibliography{references}

\begin{thebibliography}{10}
\providecommand{\url}[1]{#1}
\csname url@samestyle\endcsname
\providecommand{\newblock}{\relax}
\providecommand{\bibinfo}[2]{#2}
\providecommand{\BIBentrySTDinterwordspacing}{\spaceskip=0pt\relax}
\providecommand{\BIBentryALTinterwordstretchfactor}{4}
\providecommand{\BIBentryALTinterwordspacing}{\spaceskip=\fontdimen2\font plus
\BIBentryALTinterwordstretchfactor\fontdimen3\font minus \fontdimen4\font\relax}
\providecommand{\BIBforeignlanguage}[2]{{%
\expandafter\ifx\csname l@#1\endcsname\relax
\typeout{** WARNING: IEEEtran.bst: No hyphenation pattern has been}%
\typeout{** loaded for the language `#1'. Using the pattern for}%
\typeout{** the default language instead.}%
\else
\language=\csname l@#1\endcsname
\fi
#2}}
\providecommand{\BIBdecl}{\relax}
\BIBdecl

\bibitem{NPcompleteness}
M.~R. Gary and D.~S. Johnson, ``Computers and intractability: A guide to the theory of {NP}-{C}ompleteness,'' 1979.

\bibitem{2DSurvey}
A.~Lodi, S.~Martello, and M.~Monaci, ``Two-dimensional packing problems: A survey,'' \emph{European Journal of Operational Research}, vol. 141, no.~2, pp. 241--252, 2002.

\bibitem{genetic2D}
E.~Hopper and B.~Turton, ``A genetic algorithm for a 2{D} industrial packing problem,'' \emph{Computers \& Industrial Engineering}, vol.~37, no. 1-2, pp. 375--378, 1999.

\bibitem{Skyline}
L.~Wei, D.~Zhang, and Q.~Chen, ``A least wasted first heuristic algorithm for the rectangular packing problem,'' \emph{Computers \& Operations Research}, vol.~36, no.~5, pp. 1608--1614, 2009.

\bibitem{ShelfNextFit}
Z.~Zhu, J.~Sui, and L.~Yang, ``Bin-packing algorithms for periodic task scheduling,'' \emph{Int. Journal of Pattern Recognition and Artificial Intelligence}, vol.~25, no.~07, pp. 1147--1160, 2011.

\bibitem{3Dheuristic}
A.~Lodi, S.~Martello, and D.~Vigo, ``Heuristic algorithms for the three-dimensional bin packing problem,'' \emph{European Journal of Operational Research}, vol. 141, no.~2, pp. 410--420, 2002.

\bibitem{deeppack}
O.~Kundu, S.~Dutta, and S.~Kumar, ``Deep-{P}ack: A vision-based 2{D} online bin packing algorithm with deep reinforcement learning,'' in \emph{Proc. Int. Conf. Robot and Human Interactive Communication}, 2019, pp. 1--7.

\bibitem{solving}
H.~Hu, X.~Zhang, X.~Yan, L.~Wang, and Y.~Xu, ``Solving a new 3{D} bin packing problem with deep reinforcement learning method,'' \emph{arXiv preprint arXiv:1708.05930}, 2017.

\bibitem{generalized3D}
R.~Verma, A.~Singhal, H.~Khadilkar, A.~Basumatary, S.~Nayak, H.~V. Singh, S.~Kumar, and R.~Sinha, ``A generalized reinforcement learning algorithm for online 3{D} bin-packing,'' \emph{arXiv preprint arXiv:2007.00463}, 2020.

\bibitem{Recent3DRearrangement}
P.~Zhou, Z.~Gao, C.~Li, and N.~Y. Chong, ``An efficient deep reinforcement learning model for online 3{D} bin packing combining object rearrangement and stable placement,'' \emph{arXiv preprint arXiv:2408.09694}, 2024.

\bibitem{semi-online}
E.~G. Coffman, G.~Galambos, S.~Martello, and D.~Vigo, ``Bin packing approximation algorithms: Combinatorial analysis,'' \emph{Handbook of Combinatorial Optimization: Supplement Volume A}, pp. 151--207, 1999.

\bibitem{multidimensionalReview}
H.~I. Christensen, A.~Khan, S.~Pokutta, and P.~Tetali, ``Approximation and online algorithms for multidimensional bin packing: A survey,'' \emph{Computer Science Review}, vol.~24, pp. 63--79, 2017.

\bibitem{stopngo}
G.~Han, J.~Park, and C.~Nam, ``Stop-{N}-{G}o: Search-based conflict resolution for motion planning of multiple robotic manipulators,'' in \emph{Proc. Int. Conf. Robot. and Autom.}, 2025.

\bibitem{SynchronizedDualArm}
W.~Li, S.~Zhang, S.~Dai, H.~Huang, R.~Hu, X.~Chen, and K.~Xu, ``Synchronized dual-arm rearrangement via cooperative m{TSP},'' \emph{arXiv preprint arXiv:2403.08191}, 2024.

\bibitem{multi-agentGame}
N.~Gafur, G.~Kanagalingam, and M.~Ruskowski, ``Dynamic collision avoidance for multiple robotic manipulators based on a non-cooperative multi-agent game,'' \emph{arXiv preprint arXiv:2103.00583}, 2021.

\bibitem{ahn2022coordination}
J.~Ahn, C.~Kim, and C.~Nam, ``Coordination of two robotic manipulators for object retrieval in clutter,'' in \emph{Proc. Int. Conf. Robot. and Autom.}, 2022, pp. 1039--1045.

\bibitem{A3C}
V.~Mnih, A.~P. Badia, M.~Mirza, A.~Graves, T.~Lillicrap, T.~Harley, D.~Silver, and K.~Kavukcuoglu, ``Asynchronous methods for deep reinforcement learning,'' in \emph{Proc. Int. Conf. Machine Learning}, 2016, pp. 1928--1937.

\bibitem{Metaheuristic}
G.~Tresca, G.~Cavone, R.~Carli, A.~Cerviotti, and M.~Dotoli, ``Automating bin packing: A layer building matheuristics for cost effective logistics,'' \emph{IEEE Transactions on Automation Science and Engineering}, vol.~19, no.~3, pp. 1599--1613, 2022.

\bibitem{aaaik-BPP}
H.~Zhao, Q.~She, C.~Zhu, Y.~Yang, and K.~Xu, ``Online 3{D} bin packing with constrained deep reinforcement learning,'' in \emph{Proc. AAAI Conf. Artificial Intelligence}, vol.~35, no.~1, 2021, pp. 741--749.

\bibitem{feasible3DBPP}
H.~Zhao, C.~Zhu, X.~Xu, H.~Huang, and K.~Xu, ``Learning practically feasible policies for online 3{D} bin packing,'' \emph{Science China Information Sciences}, vol.~65, no.~1, p. 112105, 2022.

\bibitem{3Dvision}
J.~Jia, H.~Shang, and X.~Chen, ``Robot online 3{D} bin packing strategy based on deep reinforcement learning and 3{D} vision,'' in \emph{Proc. Int. Conf. Networking, Sensing and Control}, 2022, pp. 1--6.

\bibitem{Synergies}
S.~Song, S.~Yang, R.~Song, S.~Chu, W.~Zhang \emph{et~al.}, ``Towards online 3{D} bin packing: Learning synergies between packing and unpacking via {DRL},'' in \emph{Proc. Conf. Robot Learning}, 2023, pp. 1136--1145.

\bibitem{Heuritics_integratedDRL3D}
S.~Yang, S.~Song, S.~Chu, R.~Song, J.~Cheng, Y.~Li, and W.~Zhang, ``Heuristics integrated deep reinforcement learning for online 3{D} bin packing,'' \emph{IEEE Transactions on Automation Science and Engineering}, vol.~21, no.~1, pp. 939--950, 2023.

\bibitem{dual_conveyor}
Y.~Tsang, D.~Mo, K.~Chung, and C.~Lee, ``A deep reinforcement learning approach for online and concurrent 3d bin packing optimisation with bin replacement strategies,'' \emph{Computers in Industry}, vol. 164, p. 104202, 2025.

\bibitem{dual_arm_springer}
C.-Y. Weng, W.~Yin, Z.~J. Lim, and I.-M. Chen, ``A framework for robotic bin packing with a dual-arm configuration,'' in \emph{IFToMM World Congress on Mechanism and Machine Science}.\hskip 1em plus 0.5em minus 0.4em\relax Springer, 2019, pp. 2799--2808.

\bibitem{brain-inspired}
L.~Zhang, D.~Li, S.~Jia, and H.~Shao, ``Brain-inspired experience reinforcement model for bin packing in varying environments,'' \emph{IEEE Transactions on Neural Networks and Learning Systems}, vol.~33, no.~5, pp. 2168--2180, 2022.

\bibitem{lookahead}
E.~F. Grove, ``Online bin packing with lookahead,'' in \emph{Proc. ACM-SIAM Symp. Discrete Algorithms}, 1995, pp. 430--436.

\bibitem{roswebsite}
{Open Source Robotics Foundation}, ``{Robot Operating System ({ROS})},'' \url{https://www.ros.org}, [Online].

\bibitem{paszke2019pytorch}
A.~Paszke, ``Pytorch: An imperative style, high-performance deep learning library,'' \emph{arXiv preprint arXiv:1912.01703}, 2019.

\bibitem{unity_robotics_hub}
{Unity Technologies}, ``Unity robotics hub,'' \url{https://github.com/Unity-Technologies/Unity-Robotics-Hub}, [Online].

\bibitem{curobo_report23}
B.~Sundaralingam, S.~K.~S. Hari, A.~Fishman, C.~Garrett, K.~V. Wyk, V.~Blukis, A.~Millane, H.~Oleynikova, A.~Handa, F.~Ramos, N.~Ratliff, and D.~Fox, ``cu{R}obo: Parallelized collision-free minimum-jerk robot motion generation,'' 2023.

\bibitem{agilex_piper}
{AgileX Robotics}, ``{PiPER},'' \url{https://global.agilex.ai/products/piper}, [Online].

\end{thebibliography}

\end{document}